\documentclass[10pt,twocolumn,letterpaper]{article}
\usepackage[accsupp]{axessibility} 
\usepackage[pagenumbers]{cvpr} 

\usepackage[dvipsnames]{xcolor}

\definecolor{cvprblue}{rgb}{0.21,0.49,0.74}
\usepackage[pagebackref,breaklinks,colorlinks,citecolor=cvprblue]{hyperref}

\usepackage{times}  
\usepackage{helvet}  
\usepackage{courier}  
\usepackage[hyphens]{url}  
\usepackage{graphicx} 
\usepackage{natbib}  
\usepackage{caption} 
\usepackage{algorithm}
\usepackage{algorithmic}
\usepackage{comment}
\usepackage{graphicx}
\usepackage{amsmath}
\usepackage{amssymb}
\usepackage{booktabs}
\usepackage{enumitem}
\usepackage{makecell}
\usepackage{xcolor}
\usepackage{url}
\newcommand\crule[3][black]{\textcolor{#1}{\rule{#2}{#3}}}
\usepackage{booktabs, multirow} 
\definecolor{roadcolor}{RGB}{234,51,246}
\definecolor{sidewalkcolor}{RGB}{68,8,72}
\definecolor{parkingcolor}{RGB}{241,156,249}
\definecolor{othergroundcolor}{RGB}{160,32,76}
\definecolor{buildingcolor}{RGB}{246,202,69}
\definecolor{carcolor}{RGB}{111,149,238}
\definecolor{truckcolor}{RGB}{74,32,172}
\definecolor{bicyclecolor}{RGB}{136,227,242}
\definecolor{motorcyclecolor}{RGB}{37,59,146}
\definecolor{othervehiclecolor}{RGB}{96,81,242}
\definecolor{vegetationcolor}{RGB}{79, 173, 50}
\definecolor{trunkcolor}{RGB}{126, 65, 22}
\definecolor{terraincolor}{RGB}{171, 238, 105}
\definecolor{personcolor}{RGB}{234, 60, 49}
\definecolor{bicyclistcolor}{RGB}{234, 66, 195}
\definecolor{motorcyclistcolor}{RGB}{138, 42, 90}
\definecolor{fencecolor}{RGB}{238, 128, 69}
\definecolor{polecolor}{RGB}{252, 241, 161}
\definecolor{trafficsigncolor}{RGB}{233, 51, 35}
\definecolor{color1}{RGB}{176, 36, 24}
\definecolor{color2}{RGB}{119,185,0}
\definecolor{color3}{RGB}{0, 0, 200}
\definecolor{colorofteaser}{RGB}{176, 36, 24}
\definecolor{color4}{RGB}{0, 0, 0}

\newcommand{\cmark}{\ding{51}}%
\usepackage{pifont}

\newcommand{\tbblue}[1]{\textbf{\textcolor{color3}{#1}}}

\newcommand{\pub}[1]{{\color{gray}{\tiny{[{#1}]}}}}
\definecolor{mygreen}{RGB}{93,173,85}

\def\paperID{2503} 
\def\confName{CVPR}
\def\confYear{2024}

\title{\textit{Not All Voxels Are Equal}: Hardness-Aware Semantic Scene Completion \\ with Self-Distillation }

\author{Song Wang$^1$, \ \ \ Jiawei Yu$^1$, \ \ \ Wentong Li$^1$, \ \ \  Wenyu Liu$^1$,  \ \ \ Xiaolu Liu$^1$, \\ Junbo Chen$^{2}\footnotemark[1]$, \ \ \ Jianke Zhu$^{1}\thanks{Corresponding authors}$\\
	$^1$Zhejiang University \ \ \
	$^2$Udeer.ai \\ 
}

\begin{document}
\maketitle

\begin{abstract}
Semantic scene completion, also known as semantic occupancy prediction, can provide dense geometric and semantic information for autonomous vehicles, which attracts the increasing attention of both academia and industry. Unfortunately, existing methods usually formulate this task as a voxel-wise classification problem and treat each voxel equally in 3D space during training. As the hard voxels have not been paid enough attention, the performance in some challenging regions is limited. The 3D dense space typically contains a large number of empty voxels, which are easy to learn but require amounts of computation due to handling all the voxels uniformly for the existing models. Furthermore, the voxels in the boundary region are more challenging to differentiate than those in the interior. In this paper, we propose \textbf{\texttt{HASSC}} approach to train the semantic scene completion model with hardness-aware design. The global hardness from the network optimization process is defined for dynamical hard voxel selection. Then, the local hardness with geometric anisotropy is adopted for voxel-wise refinement. Besides, self-distillation strategy is introduced to make training process stable and consistent. Extensive experiments show that our \textbf{\texttt{HASSC}} scheme can effectively promote the accuracy of the baseline model without incurring the extra inference cost. 
Source code is available at: \href{https://github.com/songw-zju/HASSC}{https://github.com/songw-zju/HASSC}.
\end{abstract}

\vspace{-2mm}

\section{Introduction}
\label{sec:intro}
The accurate 3D perception of the surrounding environment is critical for both autonomous vehicles and robots~\cite{song2017semantic, zhu2021cylindrical, roldao20223d, wang2022meta}.
Early semantic scene completion works mainly focus on indoor scenes~\cite{liu2018see, li2020attention, cai2021semantic, tang2022not}. 
For outdoor driving scenarios, SemanticKITTI~\cite{behley2019semantickitti} provides the first large benchmark, in which LiDAR-based methods~\cite{roldao2020lmscnet, yan2021sparse, cheng2021s3cnet, xia2023scpnet} occupy a dominant position with promising performance. Recently, vision-centric methods~\cite{huang2021bevdet, li2022bevformer, li2023bevdepth} have made encouraging progress in bird's-eye-view (BEV) perception. Researchers have further started to complete the entire 3D semantic scene with only the camera as input and obtain impressive results~\cite{cao2022monoscene, huang2023tri, li2023voxformer}.

\begin{figure}
\centering
\includegraphics[width=0.95 \linewidth]{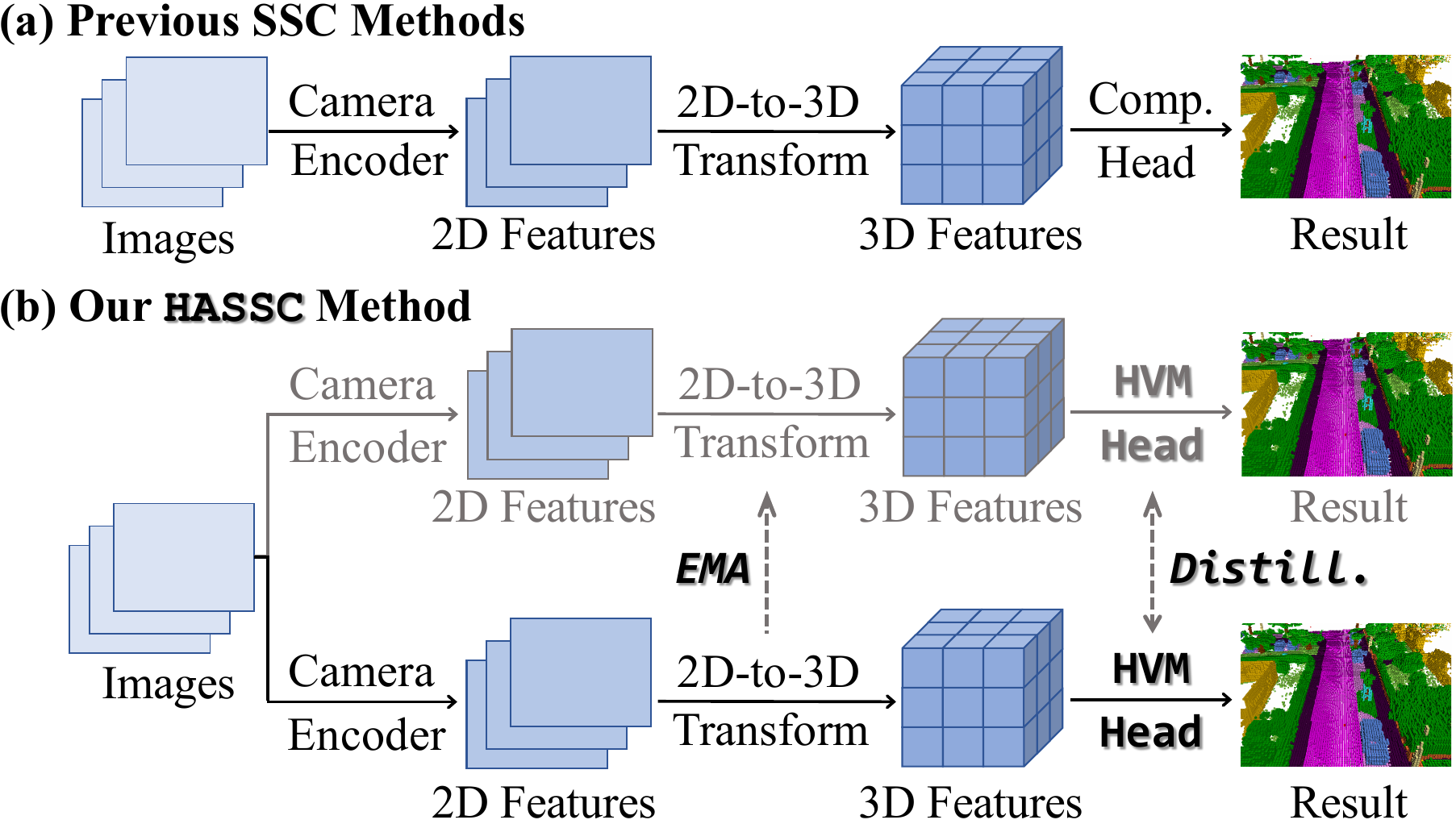}
\vspace{-2mm}
    \caption{Comparing our proposed hardness-aware semantic scene completion (\textbf{\texttt{HASSC}}) approach against previous semantic scene completion methods. We present an effective hard voxel mining (HVM) head with self-distillation during training.} 
    \label{fig:intro}
\vspace{-6mm}
\end{figure}

Generally, it is not trivial to infer semantic occupancy information in 3D dense space from the current camera observation alone. For vision-centric semantic scene completion models, the 2D backbone needs high-resolution images as input and consumes most of the GPU memory, as shown in Fig.~\ref{fig:intro}(a). 
After the forward or backward 2D-to-3D transformation~\cite{li2023fb, li2023voxformer, li2023stereoscene}, the 3D dense voxel volume features are captured while it is impossible to perform feature extraction under high-resolution in 3D space due to the GPU memory limitation. 
The 3D backbone adopts  convolution~\cite{cao2022monoscene, li2023stereoscene} or self-attention~\cite{li2023voxformer} operators and extracts fine-grained features on reduced-resolution 3D feature maps. 
As voxel-wise outputs are required at full resolution, the completion head typically derives the final result through trilinear interpolation or transposed convolution directly without considering the hardness for different voxels. 
Meanwhile, existing methods~\cite{cao2022monoscene,li2023voxformer} mainly formulate semantic scene completion as a voxel-wise classification problem and compute the loss for each voxel equally. 
Such a scheme ignores that the hardness in classifying various voxels in 3D space is quite different. 
Since more than 90\% of the voxel space is empty, these empty voxels are easy to predict but require a large amount of computation during training. 
Moreover, voxels inside an object exhibit greater predictability than those located on the boundary.

Building upon the success of hard sample mining in 2D dense prediction like object detection~\cite{shrivastava2016training, lin2017focal} and semantic segmentation~\cite{li2017not, kirillov2020pointrend, xiao2023not}, we are motivated to design a hard voxel mining strategy in 3D dense space. 
The difference between 2D pixel space and 3D voxel space is not only the extra computational cost due to dimension upgrading, but also a large number of empty voxels consuming most of memory and computation in 3D voxel space. To alleviate this problem, we propose the hard voxel mining (HVM) head with self-distillation, which select hard voxels via the global hardness and refine them with local hardness, as illustrated in Fig.~\ref{fig:intro}(b). 
Specifically, the global hardness is based on the uncertainty in predicting each voxel so that we can update the selected voxels dynamically.
As most of the voxels selected in such way are empty at the early stages of training, local hardness based on geometric anisotropy is introduced to weight the losses of different voxels. The local geometric anisotropy (LGA) of a voxel is defined as the semantic difference from its neighbors. 
We adopt the linear mapping of LGA as the local hardness to weight the voxel-wise losses from the hard area and refine their predictions.
Furthermore, self-distillation training is introduced to make the model outputs more stable and consistent. 
The teacher model is optimized by the exponential moving average (EMA) of the student model without extra training process. 
Our presented self-distillation approach works well with the hard voxel mining head to jointly improve the completion performance.

The main contributions of this work are summarized as follows:

\begin{itemize}
    \item We propose a hardness-aware semantic scene completion (\textbf{\texttt{HASSC}}) scheme that can be easily integrated into existing models without incurring extra cost for inference.
    \item We take advantage of both the global and local hardness to find the hard voxels so that their predictions can be refined by weighted voxel-wise losses during training. 
    \item A self-distillation training strategy is introduced to improve semantic scene completion through an end-to-end training manner.
    \item Extensive experiments are conducted to demonstrate the effectiveness of our presented method.  

\end{itemize}

\section{Related Work}
\label{sec:related_work}

\noindent \textbf{Semantic Scene Completion.}
The methods for outdoor semantic scene completion (SSC) can be divided into two categories according to their input:
1) \textit{LiDAR-based methods}. The grid-based approaches~\cite{roldao2020lmscnet, wilson2022motionsc} employ the occupancy grid voxelized from the sparse LiDAR point cloud as input and achieve fast inference performance with lite-weight backbone. 
The point-based methods~\cite{yan2021sparse, cheng2021s3cnet} integrate the point-wise features within the voxel space and improve the model accuracy.
Xia~\textit{et al.}~\cite{xia2023scpnet} redesign the completion network architecture and obtain the optimal accuracy with 3D input.
2) \textit{Camera-based methods.}
MonoScene~\cite{cao2022monoscene} is the pioneer work, which explores the SSC with monocular camera image firstly.
The subsequent works construct a tri-perspective view plane~\cite{huang2023tri} or design dual-path transformer decoder~\cite{zhang2023occformer} to improve the performance with single image.
VoxFormer~\cite{li2023voxformer} estimates the coarse geometry with stereo images first and obtains the non-empty proposals to perform deformable cross-attention~\cite{zhu2020deformable} on single or multiple monocular images. 
Another method category utilizes implicit representations rather than voxel-based modeling, indicating the capability for extending SSC with both LiDAR~\cite{rist2021semantic, li2023lode} and camera~\cite{ hayler2024s4c}.
Since the camera is much cheaper with greater application potentials than LiDAR, we mainly focus on vision-centric methods in this paper.

\noindent \textbf{Hard Sample Mining for Dense Prediction.}
Hard sample mining is firstly explored in 2D object detection~\cite{shrivastava2016training, lin2017focal} as the difficulty in detecting distinct objects from an image is quite different.
In 2D image segmentation, Li~\textit{et al.}~\cite{li2017not} propose a layer cascade method to segment regions with different hardness. 
Yin~\textit{et al.}~\cite{yin2019online} extract hard regions according to the loss values and re-train these areas for better performance.
Kirillov~\textit{et al.}~\cite{kirillov2020pointrend} start from the common ground of image rendering and segmentation, and refine the edge regions of objects in the image from the feature level. Deng~\textit{et al.}~\cite{deng2022nightlab} use an auxiliary detection network to find hard areas at nighttime and conduct segmentation refinement in both training and inference. 
Xiao~\textit{et al.}~\cite{xiao2023not} propose a pixel hardness learning method by making use of global and historical loss values. 
The above methods are all designed for images in 2D pixel space. 
Li~\textit{et al.}~\cite{li2019depth} propose local geometric anisotropy to weight voxel-wise cross-entropy losses, which does not perform well in outdoor scenes. 
In this paper, we propose to conduct effective hard sample mining in 3D dense voxel space for driving scenes.

\begin{figure*}[t]
\centering
\includegraphics[width=1.0\linewidth]{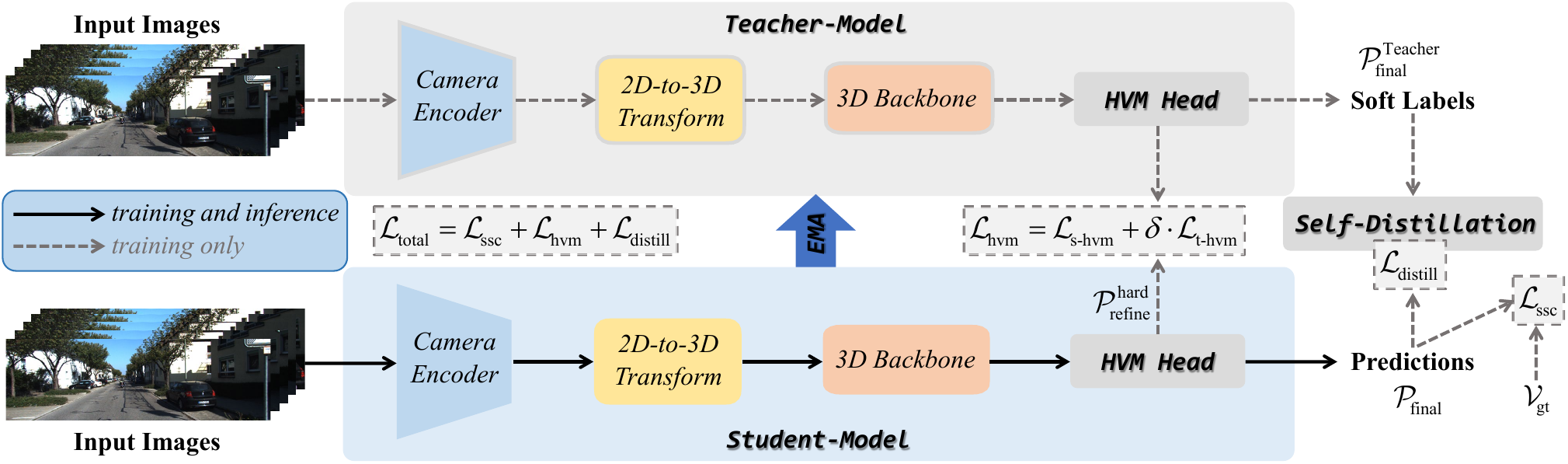}
    \caption{Overview of \textbf{Hardness-Aware Semantic Scene Completion (\textbf{\texttt{HASSC}})} pipeline. We take the camera images as input and construct 3D feature volume by \textit{Camera Encoder} and \textit{2D-to-3D Transform}. With the fine-grained features provided by \textit{3D Backbone}, we propose hard voxel mining (HVM) head to make the model concentrate on hard voxels. The teacher-model has the same architecture as student-model, which is updated by the exponential moving average (EMA) of student. The stable predictions can be achieved by taking advantage of both the self-distillation and HVM head.}
    \label{fig:framework}
    \vspace{-2mm}
\end{figure*}

\noindent \textbf{Self-Distillation.} 
Knowledge distillation is firstly proposed to learn dark knowledge from well-trained large models for model compression~\cite{buciluǎ2006model, hinton2015distilling}. 
A series of subsequent works have been presented to improve the learning efficiency and capability of student models~\cite{furlanello2018born, heo2019comprehensive, yang2022masked, zhao2022decoupled}. 
In autonomous driving, knowledge distillation, especially cross-modality distillation, has shown its great potential in improving model accuracy~\cite{chen2022bevdistill, zhou2023unidistill,wang2023lidar2map} and compressing models~\cite{yang2022towards, cho2023itkd, zeng2023distilling}.
However, these methods usually need to train a stronger teacher with more parameters or other modality at first, which incur additional training costs. 
Inspired by successful applications of self-distillation in other fields including 2D and 3D semantic segmentation~\cite{zhang2019your, ji2021refine, lan2021discobox, li2022self}, we introduce self-distillation training strategy for semantic scene completion without extra designed models.

\section{Method}
\label{sec:method}
The semantic scene completion task predicts a dense semantic voxel volume $\mathcal{V} \in  \mathbb{R}^{X \times Y \times Z}$ in front of the vehicle with only the observation from onboard sensors including camera and LiDAR. 
($X$, $Y$, $Z$) represent the length, width and height of the 3D volume, respectively.
Each voxel in $\mathcal{V}$ is either empty or occupied with one semantic class. We mainly consider the more challenging 2D input for its low cost and great application potentials.

\subsection{Overview}
In this work, we aim to provide a hard sample mining solution in voxel space for 3D dense prediction. The overall pipeline of Hardness-Aware Semantic Scene Completion (\textbf{\texttt{HASSC}}) is illustrated in Fig.~\ref{fig:framework}. 
Our proposed hard voxel mining (HVM) head and self-distillation training strategy are independent from specific network, which can be easily integrated into off-the-shelf methods.

The typical semantic scene completion network, \textit{e.g.}, MonoScene~\cite{cao2022monoscene} and VoxFormer~\cite{li2023voxformer}, consists of camera encoder, 2D-to-3D transformation, 3D backbone and completion head.

\noindent \textbf{Camera Encoder.} The camera encoder is made of an image backbone and a neck, which extracts semantic and geometric feature $\mathcal{F}^{\text{2D}} \in  \mathbb{R}^{H' \times W' \times D'} $ from images under the perspective view. $H' \times W'$ is the 2D feature resolution, and $D'$ is the dimension. The extracted feature is the basis to construct 3D voxel volume in the following.

\noindent \textbf{2D-to-3D Transformation.} The 2D-to-3D transformation in semantic scene completion is similar to view transformation in BEV perception, which can be divided into two paradigms, including forward projection~\cite{philion2020lift, huang2021bevdet, li2023stereoscene} and backward projection~\cite{wang2022detr3d, li2022bevformer, li2023voxformer}. 
With the extracted image feature $\mathcal{F}^{\text{2D}}$, we construct 3D volume feature $\mathcal{F}^{\text{3D}} \in  \mathbb{R}^{ X' \times Y' \times Z' \times D}$ by explicit geometry estimation and query-based 3D-to-2D back projection~\cite{li2023voxformer}. To reduce memory consumption and computational cost, $X' \times Y' \times Z'$ as the 3D feature resolution is smaller than $X \times Y \times Z$. $D$ is the 3D feature dimension.

\noindent \textbf{3D Backbone.} The 3D Backbone performs self-attention~\cite{li2023voxformer} or convolution~\cite{cao2022monoscene, li2023stereoscene} on the voxel volume feature $\mathcal{F}^{\text{3D}} $ from the 2D-to-3D Transformation and obtain fine-grained feature $\mathcal{F}_{\text{fine}}^{\text{3D}} \in  \mathbb{R}^{X' \times Y' \times Z' \times D}$. Moreover, the completion head uses $\mathcal{F}_{\text{fine}}^{\text{3D}}$ to obtain the final prediction $\mathcal{P}_{\text{final}} \in \mathbb{R}^{X \times Y \times Z \times C}$, where $C$ is the number of total classes including empty and semantic categories.

\subsection{Hard Voxel Mining Head}
\begin{figure}
\centering
\includegraphics[width=0.95 \linewidth]{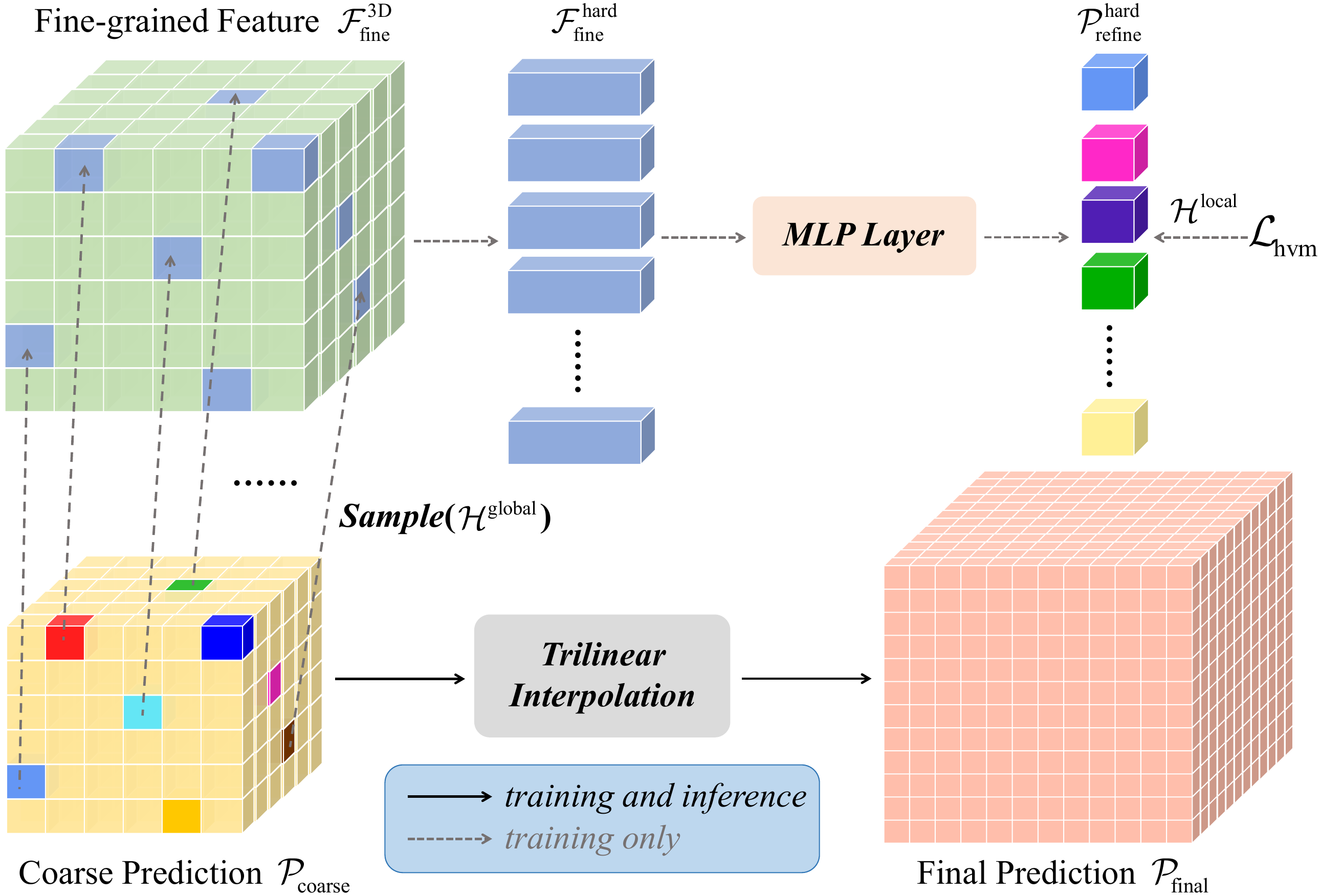}   
\caption{Illustration of \textbf{Hard Voxel Mining (HVM) Head}. 
At training stage, $N$ hard voxels are selected with respect to their global hardness and random sampling. Then, we re-sample the corresponding fine-grained features and employ \textit{MLP Layer} to refine their predictions, which are supervised by the ground truth and local hardness. For inference, we directly utilize \textit{Trilinear Interpolation} to obtain final prediction.}
\label{fig:hvm}
    \vspace{-4mm}
\end{figure}

The completion head in SSC usually up-samples the fine-grained feature $\mathcal{F}_{\text{fine}}^{\text{3D}}$ from the 3D backbone through trilinear interpolation or transposed convolution to obtain the completion result $\mathcal{P}_{\text{final}}$ of full-resolution. Since this process does not take into account the hardness of each voxel, the performance is poor in some difficult regions. 
Our proposed hard voxel mining (HVM) head is based on the vanilla completion head and selects hard voxels during training to refine their predictions.
In the following, we firstly introduce the definitions of global hardness and local hardness, and then explain the working flow of proposed HVM head, as illustrated in Fig.~\ref{fig:hvm}.

\noindent \textbf{Global Hardness.}
With the fine-grained feature $\mathcal{F}_{\text{fine}}^{\text{3D}}$, we obtain the coarse prediction $\mathcal{P}_{\text{coarse}} \in  \mathbb{R}^{ X' \times Y' \times Z' \times C}$ by the single layer Multi-Layer Perceptron (MLP) and \textit{softmax} function firstly.  Let ($i,j,k$) denote the voxel index. For the prediction of each voxel $p_{(i,j,k)} \in \mathbb{R}^{1 \times C} $ in $\mathcal{P}_{\text{coarse}}$ , we rank the probabilities of each class $\{p^{1}, p^{2}, ..., p^{C} \}$ in decreasing order. 
The largest probability in $C$ classes is represented as ${p^a}$, and the second largest one is denoted as ${p^b}$.
Then, the global hardness $\mathcal{H}_{{i,j,k}}^{\text{global}}$ of this voxel is defined as follows
\begin{equation}
\mathcal{H}_{{i,j,k}}^{\text{global}}=\frac{1}{p^a-p^b}.
\end{equation}

We obtain $\mathcal{H}^{\text{global}} \in \mathbb{R}^{ X' \times Y' \times Z'}$ by computing all the predictions of voxels in $\mathcal{P}_{\text{coarse}}$.
$\mathcal{H}^{\text{global}}$ measures the uncertainty of the semantic scene completion prediction between the class $a$ and $b$, which varies with the optimization of the network. 
The value of $\mathcal{H}^{\text{global}}$ indicates the global hardness of predicting a certain voxel during the overall training process. In this paper, we mainly employ $\mathcal{H}^{\text{global}}$ to select hard voxels and refine their predictions.

\begin{figure}
\centering
\includegraphics[width=1.0 \linewidth]{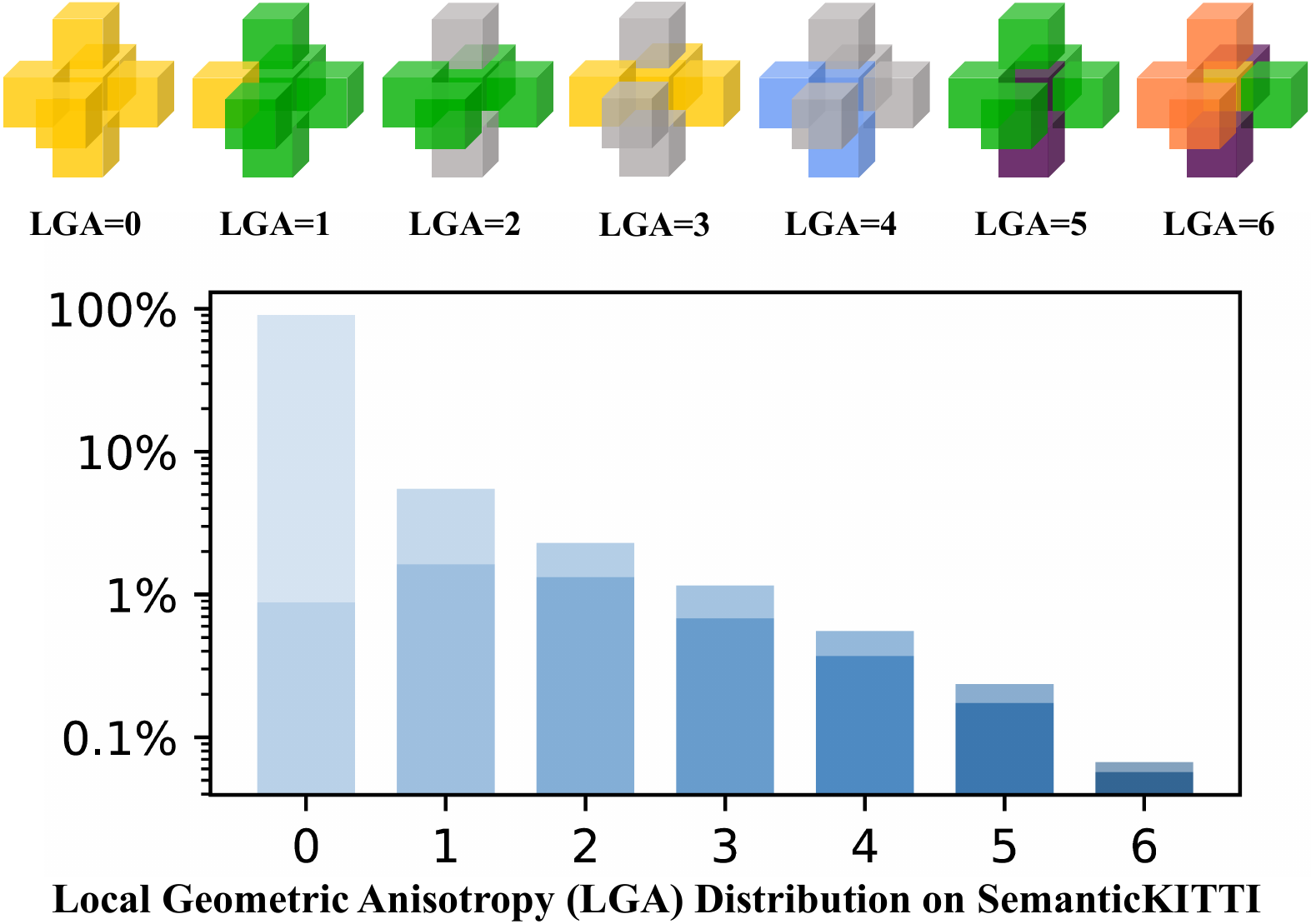}
\caption{Illustration of \textbf{Local Geometric Anisotropy (LGA)}. 
The upper figure gives the examples of different LGA values, which are all from real scenarios.
The lower figure shows the distribution of LGA values in SemanticKITTI~\cite{behley2019semantickitti}.
We use dark and light colors to represent the proportion of non-empty and empty voxels in each LGA value category, respectively.}
\label{fig:lga}
    \vspace{-4mm}
\end{figure}

\noindent \textbf{Local Hardness.}
In 3D dense space, each voxel involves different geometric information, which depends on its various location. 
Since the voxels in the boundary region pose a greater challenge than those in the interior,
the local information is crucial for instructing the model to find genuinely hard voxels.
We adopt local geometric anisotropy~\cite{li2019depth} (LGA) $\mathcal{A} \in  \mathbb{R}^{X \times Y \times Z}$ as the basis of the local hardness $\mathcal{H}^{\text{local}} \in  \mathbb{R}^{X \times Y \times Z}$ on the selected voxels. For each voxel $v_{(i,j,k)}$ in $\mathcal{V}$, the LGA $\mathcal{A}_{i,j,k}$ is computed with its neighbors $\{ v^1, v^2,.., v^M \}$ at $M$ different directions
\begin{equation}
\mathcal{A}_{i,j,k}=\sum_{m=1}^M\left(v_{\text{gt}} \oplus v_{\text{gt}}^{m}\right), 
\label{eq:lga}
\end{equation}
where $v_{\text{gt}}$ and $v_{\text{gt}}^{m}$ are the semantic labels of $v$ and $v^{m}$ ($m=1,...,M$), respectively. $\oplus$ denotes the exclusive disjunction operation. It returns $0$ or $1$ when $v$ and $v^m$ have the same semantic label or not. Note that we calculate the LGA of all the voxels including empty ones. 
In our implementation, we set $M$ to $6$ and compute the voxels in the up/down, front/back, and left/right directions.
The examples and distribution information on SemanticKITTI~\cite{behley2019semantickitti} for different LGA values are provided in Fig.~\ref{fig:lga}.

Then, the corresponding local hardness $\mathcal{H}_{{i,j.k}}$ is defined as below
\begin{equation}
\mathcal{H}_{{i,j,k}}^{\text{local}}=\alpha+ \beta \mathcal{A}_{i,j,k},
\label{eq:loc_hard}
\end{equation}
where $\alpha$ and $\beta$ are the coefficients linearly mapping $\mathcal{A}$ onto $\mathcal{H}^{\text{local}}$.
$\mathcal{H}^{\text{local}}$ measures the semantic difference between the selected voxel and its neighbors.
The value of $\mathcal{H}^{\text{local}}$ reflects the geometric position on the object of the voxel. 
{We adopt local hardness $\mathcal{H}^{\text{local}}$ to weight the selected voxels and make the model focus on more challenging voxel positions.}

\noindent \textbf{Hard Voxel Selection.}
During training, we select $N$ hard voxels from the coarse prediction $\mathcal{P}_{\text{coarse}}$ with the global hardness $\mathcal{H}^{\text{global}}$.
Since $N \ll X' \times Y' \times Z'$, selecting $N$ voxels with the largest $\mathcal{H}^{\text{global}}$ in $\mathcal{P}_{\text{coarse}}$ directly  may cause the SSC network to fall into over-fitting in the local area at the beginning.
Motivated by PointRend~\cite{kirillov2020pointrend}, we firstly over-generate proposal voxels by randomly sampling $tN$ voxels ($t > 1$) with a homogeneous distribution in 3D dense space. Secondly, $\omega N$ hard voxels ($\omega \in [0,1]$) are selected from $tN$ proposals by sorting the global hardness, which is calculated from the corresponding coarse prediction region $\mathcal{P}_{\text{coarse}}$. 
Thirdly, remaining $(1 - \omega)N$ voxels are randomly sampled from 3D space to prevent over-fitting in training. 
Finally, we obtain $N$ hard voxels with the coordinates $\mathcal{V}^{\text{hard}} \in \mathbb{R}^{N \times 3}$.

\noindent \textbf{Voxel-wise Refinement Module.} Given $N$ selected hard voxels, we re-sample their corresponding features from the fine-grained 3D volume features $\mathcal{F}^{\text{3D}}_{\text{fine}}$ by the coordinates $\mathcal{V}^{\text{hard}}$ 
and obtain $\mathcal{F}^{\text{hard}}_{\text{fine}} \in \mathbb{R}^{N \times D}$. 
Then, the hard voxels are refined with a lightweight network consisting of MLPs like PointNet~\cite{qi2017pointnet}, since the voxel-wise prediction can be considered as point-wise segmentation problem as below
\begin{equation}
\mathcal{P}_{\text{refine}}^{\text{hard}}=\text{MLP}(\mathcal{F}^{\text{hard}}_{\text{fine}}),
\end{equation}
where $\mathcal{P}_{\text{refine}}^{\text{hard}}$ is the refined prediction of the selected hard voxels.

In training process, we need to re-sample the corresponding semantic labels of hard voxels from the ground truth. 
As the majority of 3D dense space is empty, we observe that the selected hard voxels are also predominantly empty ones at the beginning. 
In fact, part of these empty voxels are not inherently challenging to distinguish. 
If we treat the selected hard voxels equally, it may cause the SSC network to focus on some samples that are not actually difficult in the early stages of training.
Therefore, we adopt the predefined local hardness $\mathcal{H}^{\text{local}}$ to weight the selected hard voxels and make the SSC network with the auxiliary MLP-based refinement head concentrate on the harder samples at local position. 
Specifically, the local hardness of selected hard voxels is computed by Eq.~\ref{eq:lga} and Eq.~\ref{eq:loc_hard}. Then, the hard voxel mining loss is calculated as follows

\begin{equation}
\mathcal{L}_{\text{s-hvm}}=\frac{1}{N} \sum_{n=1}^N  \mathcal{H}^{\text{local}}_{n} \cdot \mathrm{CE}({v}_{\text{refine}}^{n}, v_{\text{gt}}^{n}),
\label{eq:hvm}
\end{equation}
where $\mathrm{CE}(\cdot, \cdot)$ denotes the cross entropy loss function. ${v}_{\text{refine}}^{n}$ and $v_{\text{gt}}^{n}$ are the refined prediction and the semantic label of the selected $n$-th hard voxel, respectively.

At inference stage, we directly use trilinear interpolation to obtain the final completion result $\mathcal{P}_{\text{final}}$ without introducing extra computational burden, as illustrated in Fig.~\ref{fig:hvm}.

\subsection{Self-Distillation}
To further train a robust model with higher performance without extra well-trained teacher model, we propose to perform self-distillation with the hard voxel mining head. 
The teacher model in our \textbf{\texttt{HASSC}} scheme has the same network architecture as the student model, as shown in Fig.~\ref{fig:framework}. Following~\cite{tarvainen2017mean, he2020momentum}, we construct a mean teacher by exponential moving average (EMA) to achieve better stability and consistency between iterations. 
During training process, the parameters of teacher model share the same initialization as student at step $0$ and update at next steps as below
\begin{equation}
\theta_{t+1}^{\text{Teacher}}=\gamma \theta_{t}^{\text{Teacher}}+(1-\gamma) \theta_{t+1}^{\text{Student}},
\label{eq:ema}
\end{equation}
where $\gamma =\min \left(1-\frac{1}{t+1}, 0.99\right)$. $\theta_t^{\text{Teacher}}$ and $\theta_t^{\text{Student}}$ are the learnable parameters of teacher and student model at step $t$. We only optimize parameters of the student model $\theta^{\text{Student}}$ and update the teacher network by Eq.~\ref{eq:ema}. 

Since the teacher network has the same hard voxel mining head as student, $N$ hard voxels $\mathcal{V}^{\text{hard-T}} \in \mathbb{R}^{N \times 3}$ can be obtained during training. We use $\mathcal{V}^{\text{hard-T}}$ to sample the final result $\mathcal{P}_{\text{final}}$ from the student branch. 
Then, the teacher-guided hard voxel mining loss $\mathcal{L}_{\text{t-hvm}}$ is computed by Eq.~\ref{eq:hvm} with the corresponding local hardness, final prediction and ground truth.
In case of the large number of voxels, we employ $\mathcal{L}_{\text{t-hvm}}$ to make the hard voxel selection by HVM head in student model more stable and consistent.

Moreover, we adopt Kullback–Leibler divergence ($\boldsymbol{D}_{\mathrm{KL}}$) to instruct student model to learn from the online soft labels $\mathcal{P}_{\text {final}}^{\text {Teacher}}$ provided by teacher-branch as follows 
\begin{equation}
\mathcal{L}_{\text{distill}} =\lambda e^{\mu} \cdot \boldsymbol{D}_{\mathrm{KL}}\left(\mathcal{P}_{\text {final}}^{\text {Teacher}} \| \mathcal{P}_{\text {final}}\right), 
\label{eq:kld}
\end{equation}
where $\lambda$ is the weight coefficient for distillation. 
$\mu \in [0, 1]$ is the mean intersection over union (mIoU) value between the prediction of the current frame with teacher model $\mathcal{P}_{\text {fine}}^{\text {Teacher}}$ and the corresponding ground truth $\mathcal{V}_{\mathrm{gt}}$.

\subsection{Training and Inference}
\noindent \textbf{Overall Loss Function for Training.} 
In this work, we follow the common settings  ~\cite{cao2022monoscene, li2023voxformer} and treat the semantic scene completion (SSC) as a voxel-wise classification problem. 
Overall, the total training loss of our proposed \textbf{\texttt{HASSC}} is composed of three terms as below
\begin{equation}
\mathcal{L}_{\text{total}}=\mathcal{L}_{\text{ssc}}+\mathcal{L}_{\text{hvm}}+\mathcal{L}_{\text{distill}},
\end{equation}
where $\mathcal{L}_{\text{ssc}}$ is the commonly used loss for SSC. $\mathcal{L}_{\text{ssc}}$ consists of  
weighted cross entropy loss $\mathcal{L}_{\text{wce}}$ and scene-class affinity losses as follows
\begin{equation}
\mathcal{L}_{\text{ssc}}=\mathcal{L}_{\text{wce}}+\mathcal{L}_{\text{sem}} + \mathcal{L}_{\text{geo}},
\end{equation}
where $\mathcal{L}_{\text{sem}}$ and $\mathcal{L}_{\text{geo}}$ are scene-class affinity losses optimized for semantics and geometry, respectively.

Additionally, the hard voxel mining loss $\mathcal{L}_{\text{hvm}}$ is made of $\mathcal{L}_{\text{s-hvm}}$ and $\mathcal{L}_{\text{t-hvm}}$:
\begin{equation}
\mathcal{L}_{\text{hvm}}=\mathcal{L}_{\text{s-hvm}}+\delta \cdot \mathcal{L}_{\text{t-hvm}},
\end{equation}
where $\delta$ is the trade-off coefficient between student and teacher.

\noindent \textbf{Inference.} The student-branch is well optimized during training, which not only digs out hard samples and refine them with fine-grained features but also makes use of the soft labels provided by the teacher-branch. During the inference process, we only need to preserve the student-branch without incurring the extra computational cost.

\begin{table*}[!htb]\centering
\renewcommand\tabcolsep{2.3pt}
\footnotesize
\begin{tabular}{l|ccc|ccc|ccc|ccc|ccc|ccc}\toprule
\textbf{Methods} 
&
\multicolumn{3}{c|}{\textbf{VoxFormer-S}~\cite{li2023voxformer}} 
&
\multicolumn{3}{c|}
{\textbf{\textbf{\makecell[c]{\texttt{HASSC} \\ VoxFormer-S}}}} 
&
\multicolumn{3}{c|}{\textbf{VoxFormer-T}~\cite{li2023voxformer}} 
&
\multicolumn{3}{c|}{\textbf{\makecell[c]{\texttt{HASSC} \\ VoxFormer-T}}} 
&
\multicolumn{3}{c|}{\textbf{StereoScene$^{\dag}$}~\cite{li2023stereoscene}} 
&
\multicolumn{3}{c}{\textbf{\makecell[c]{\texttt{HASSC} \\ StereoScene}}} 
\\\midrule
\textbf{Modality} &\multicolumn{3}{c|}{\textbf{Camera}} &\multicolumn{3}{c|}{\textbf{Camera}} &\multicolumn{3}{c|}{\textbf{Camera}} &\multicolumn{3}{c|}{\textbf{Camera}} &\multicolumn{3}{c|}{\textbf{Camera}} &\multicolumn{3}{c}{\textbf{Camera}} 
\\\midrule
\textbf{Range} &\textbf{S} &\textbf{M} &\textbf{L} &\textbf{S} &\textbf{M} &\textbf{L}  &\textbf{S} &\textbf{M} &\textbf{L}  &\textbf{S} &\textbf{M} &\textbf{L}  &\textbf{S} &\textbf{M} &\textbf{L}  &\textbf{S} &\textbf{M} &\textbf{L} \\\midrule
\textbf{IoU (\%)}$\uparrow$
& 65.35 & 57.54 & 44.02 
& \tbblue{65.54} & \tbblue{57.99} & \tbblue{44.82}
& {65.38} & {{57.69}} & {{44.15}}
& \tbblue{66.05} & \tbblue{58.01} & \tbblue{44.58} 
& 65.70 & 56.84 & 43.66
& {65.52} & \tbblue{57.01} & \tbblue{44.55}\\
\textbf{mIoU (\%)}$\uparrow$ 
& 17.66 & 16.48 & 12.35
& \tbblue{18.98} & \tbblue{17.95} & \tbblue{13.48}  
& {21.55} & {18.42} & {13.35} 
& \tbblue{24.10} & \tbblue{20.27} & \tbblue{14.74} 
&23.27 &21.15 &{15.24} 
&\tbblue{24.43} & \tbblue{22.17} & \tbblue{{15.88}} \\  
\midrule
\crule[carcolor]{0.13cm}{0.13cm} \textbf{car} (3.92\%) 
& 39.78 & 35.24 & 25.79 
& \tbblue{42.37} & \tbblue{36.78}  & \tbblue{27.23} 
& {44.90} & {37.46} & {26.54}
& \tbblue{45.79} & \tbblue{37.70} & \tbblue{27.33} 
& 47.05 & 43.52 & {31.15}  & 46.47  & 43.02   & 30.64   \\
\crule[bicyclecolor]{0.13cm}{0.13cm} \textbf{bicycle} (0.03\%) 
& 3.04 & 1.48 & 0.59 
& 2.72 & \tbblue{2.26} & \tbblue{0.92}  
& {5.22} & {2.87} & {1.28}
& 4.23 & 2.11 & 1.07 
&{2.38} &{2.15} &{1.05}   & \tbblue{4.20}  & \tbblue{2.63}   &  \tbblue{1.20}  \\
\crule[motorcyclecolor]{0.13cm}{0.13cm} \textbf{motorcycle} (0.03\%)
& 2.84 & 1.10 & 0.51 
& \tbblue{4.49} & \tbblue{1.63} & \tbblue{0.86} 
& {2.98} & {1.24} &{0.56} 
& \tbblue{5.64} & \tbblue{2.03} & \tbblue{1.14}
& {4.78} &{2.84} & {1.55}  
& \tbblue{5.26}  &  \tbblue{3.34}  &  0.91  \\
\crule[truckcolor]{0.13cm}{0.13cm} \textbf{truck} (0.16\%) 
& 7.50 & 7.47 & 5.63 
& 6.25 & \tbblue{11.00} &  \tbblue{9.91}  
& {9.80} &{10.38} &{7.26}
& \tbblue{22.89} & \tbblue{21.90} & \tbblue{17.06} 
&{18.72} & {22.48} & {17.55}  & \tbblue{24.94}  &  \tbblue{34.73}  & \tbblue{23.72}   \\
\crule[othervehiclecolor]{0.13cm}{0.13cm} \textbf{other-veh.} (0.20\%) 
& 8.71 & 4.98 & 3.77
& \tbblue{14.77} & \tbblue{8.85} &  \tbblue{5.61}  
& {17.21} & {10.61} & {7.81} 
& \tbblue{22.71} & \tbblue{13.52} & \tbblue{8.83} 
&{17.33} &{13.79} &{9.26}  
& \tbblue{20.61}  & \tbblue{14.24}   &  7.77  \\
\crule[personcolor]{0.13cm}{0.13cm} \textbf{person} (0.07\%) 
&  4.10 & 3.31 & 1.78  
&  \tbblue{5.11} & \tbblue{4.89} & \tbblue{2.80}   
& {4.44} & {3.50} &{1.93} 
& \tbblue{5.12} & \tbblue{4.18} & \tbblue{2.25} 
&{6.31} &{4.37} & {2.17}  & 6.06  & 3.58   &  1.79  \\
\crule[bicyclistcolor]{0.13cm}{0.13cm} \textbf{bicyclist} (0.07\%)
& 6.82 & 7.14 & 3.32
& \tbblue{6.87} & \tbblue{8.57} &  \tbblue{4.71}  
& {2.65} & {3.92} & {1.97} 
& \tbblue{4.09} & \tbblue{6.58} & \tbblue{4.09} 
&{7.70} &{4.75} &{2.30}   & \tbblue{8.22}  & \tbblue{5.65}   &  \tbblue{2.47}  \\
\crule[motorcyclistcolor]{0.13cm}{0.13cm} \textbf{motorcyclist} (0.05\%) & 0.00 & 0.00 & 0.00  & 0.00 & 0.00 & 0.00 &  0.00 &  0.00 &  0.00  &  0.00 & 0.00 & 0.00 &0.00 &0.00 &0.00   & 0.00  & 0.00   &   0.00 \\
\crule[roadcolor]{0.13cm}{0.13cm} \textbf{road} (15.30\%) 
 & 72.40 & 65.74 & 54.76  
& \tbblue{74.49} & \tbblue{68.04} & \tbblue{57.05} 
& {75.45} &{66.15} &{53.57} 
& \tbblue{78.51} & \tbblue{70.02} & \tbblue{57.23}
&79.24 & 74.16 & {61.86}   &  \tbblue{80.61} &  \tbblue{75.53}  & \tbblue{62.75}  \\
\crule[parkingcolor]{0.13cm}{0.13cm} \textbf{parking} (1.12\%) 
& 10.79 &  18.49 & 15.50 
& \tbblue{15.49} & \tbblue{21.23} &  \tbblue{15.90}  
& {21.01} &{23.96} &{19.69} 
& \tbblue{29.43} &  \tbblue{26.69} &\tbblue{19.89}
&{21.33} &{21.19} &{17.02}  & \tbblue{25.21}  &  \tbblue{25.95}  &  \tbblue{20.20}  \\
\crule[sidewalkcolor]{0.13cm}{0.13cm} \textbf{sidewalk} (11.13\%) 
& 39.35 & 33.20 & 26.35 
& \tbblue{42.69} & \tbblue{36.32} & \tbblue{28.25} 
& {45.39} &{34.53} & {26.52} 
& \tbblue{51.69} & \tbblue{38.83} & \tbblue{29.08} 
& 50.71 &41.86 &{30.58}   &  \tbblue{52.68} &  \tbblue{43.61}  & \tbblue{32.40}   \\
\crule[othergroundcolor]{0.13cm}{0.13cm} \textbf{other-grnd}(0.56\%) 
& 0.00 & 1.54 & 0.70 
& \tbblue{0.02} & \tbblue{2.38} &  \tbblue{1.04}  
&  0.00 &{0.76} &{0.42} 
& 0.00 & \tbblue{1.55} & \tbblue{1.26}
& {0.00} & {1.12} & {0.85}  &  0.00 &  0.18  &  0.51  \\
\crule[buildingcolor]{0.13cm}{0.13cm} \textbf{building} (14.10\%) 
& 17.91 & 24.09 &  17.65 
& \tbblue{22.78} & \tbblue{27.30} &  \tbblue{19.05}  
& {25.13} & {29.45} & {19.54} 
& \tbblue{27.99} & \tbblue{30.81} & \tbblue{20.19} 
& {26.98} & 32.52 & 22.71   &  \tbblue{29.09} & 31.68   & \tbblue{22.90}   \\
\crule[fencecolor]{0.13cm}{0.13cm} \textbf{fence} (3.90\%) 
& 12.98 & 10.63 & 7.64
& 9.81 & 8.70 &  6.58  
& {16.17} &{11.15} & {7.31}
& \tbblue{17.09} & \tbblue{11.65} & \tbblue{7.95}
&{22.50} &{14.26} &{8.73}   & 20.88  &  13.32  & 8.67   \\
\crule[vegetationcolor]{0.13cm}{0.13cm} \textbf{vegetation} (39.3\%) 
& 40.50 & 34.68 & 24.39 
& 40.49 &\tbblue{35.53} &  \tbblue{25.48}  
& {43.55} & {38.07} & {26.10} 
& \tbblue{44.68} & \tbblue{38.93} & \tbblue{27.01} 
&40.20 & 36.10 & 24.81  & \tbblue{40.29}  & \tbblue{36.44}  & \tbblue{26.27}   \\
\crule[trunkcolor]{0.13cm}{0.13cm} \textbf{trunk} (0.51\%) 
& 15.81 & 10.64 & 5.08  
& 14.93 & \tbblue{11.25} &  \tbblue{6.15}  
& {21.39}  &{12.75} & {6.10} 
& \tbblue{22.22} & \tbblue{14.11} & \tbblue{7.71} 
&21.45 & 15.28 & 7.17  &  \tbblue{21.65} & 14.92   & 7.14   \\
\crule[terraincolor]{0.13cm}{0.13cm} \textbf{terrain} (9.17\%) 
& 32.25 &  35.08 & 29.96  
& \tbblue{36.66} & \tbblue{38.28} &  \tbblue{32.94}  
& {42.82} & {39.61} & {33.06} 
& \tbblue{47.04} & \tbblue{41.37} & \tbblue{33.95} 
& 45.75 & 43.67 & 34.87  & \tbblue{48.50}  &  \tbblue{46.95}  & \tbblue{38.10}   \\
\crule[polecolor]{0.13cm}{0.13cm} \textbf{pole} (0.29\%) 
&  14.47 &  11.95 & 7.11 
& \tbblue{15.25} & \tbblue{12.48} &  \tbblue{7.68}  
&  {20.66} &{15.56} & {9.15} 
& 18.95 & 14.76 & \tbblue{9.20} 
& 20.43 & 18.95 & 10.66   &  18.67 &  16.34  &  9.00  \\
\crule[trafficsigncolor]{0.13cm}{0.13cm} \textbf{traf.-sign} (0.08\%) 
 & 6.19 & 6.29 & 4.18 
& 5.52 & 5.61 &  4.05  
& {10.63} &{8.09} & {4.94} 
& 9.89 & \tbblue{8.44} & 4.81
& 9.21 & 8.91 &{5.19}   &  \tbblue{10.88} &  \tbblue{9.08}  &  \tbblue{5.23}  \\
\bottomrule
\end{tabular}
\vspace{-2mm}
\caption{Quantitative comparisons against the selected baseline methods on the \textbf{validation set} of SemanticKITTI~\cite{behley2019semantickitti}. $\dag$ denotes the results are reproduced from the original implementation. ``S'', ``M'' and ``L'' represent the short range ($12.8 \times 12.8 \times 6.4\text{m}^{3}$), middle range ($25.6 \times 25.6 \times 6.4\text{m}^{3}$) and long/full range ($51.2 \times 51.2 \times 6.4\text{m}^{3}$), respectively. The improved results compared to the corresponding baselines are marked in \tbblue{blue}.}
\label{tab:comp_val}
\vspace{-4mm}
\end{table*}

\section{Experiments}
\label{sec:exps}
\subsection{Setup}

\noindent \textbf{Dataset.} The SemanticKITTI dataset~\cite{behley2019semantickitti} is the first large semantic scene completion benchmark for outdoor scenes, which contains LiDAR scans and front camera images from KITTI Odometry Benchmark~\cite{geiger2012we}. 
The ground truth is generated from the accumulated LiDAR semantic segmentation labels, which is represented as the $256 \times 256 \times 32$ voxel grids with a resolution of $0.2 \text{m}$. 
Each voxel grid is annotated as one of 19 semantic classes or 1 empty class. 
We adopt the same setting as in~\cite{behley2019semantickitti, geiger2012we} and split the total 22 sequences into (00-07, 09-10) / (08) / (11-21) for training/validation/test sets.

\noindent \textbf{Evaluation Metrics.}
The mean intersection over union (mIoU) on 19 semantic classes is reported to evaluate the quality of semantic scene completion (SSC). 
Moreover, we adopt intersection over union (IoU) to measure the performance of class-agnostic scene completion (SC), which reflects the 3D geometric quality with 2D camera images as input. 
Besides, we calculate the IoU and mIoU at different ranges from the ego car on the validation set, which include the volume of $12.8 \times 12.8 \times 6.4\text{m}^{3}$ (short range, S), $25.6 \times 25.6 \times 6.4\text{m}^{3}$ (middle range, M), and $51.2 \times 51.2 \times 6.4\text{m}^{3}$ (long/full range, L). In practical, the perception results at closer range are more critical to vehicle safety.

\noindent \textbf{Implementation Details.}
Our proposed \textbf{\texttt{HASSC}} method is designed as a generic training scheme to improve the performance of existing methods in hard regions. To demonstrate the efficacy of our approach, we choose the state-of-the-art methods including VoxFormer-S~\cite{li2023voxformer}, VoxFormer-T~\cite{li2023voxformer}\footnote{https://github.com/NVlabs/VoxFormer} and StereoScene~\cite{li2023stereoscene}\footnote{https://github.com/Arlo0o/StereoScene} as our baseline models. 
VoxFormer-S only adopts the current frame from left camera as input while VoxFormer-T combines the previous $4$ images.
StereoScene uses both the left and right camera images to train the model.
The input image size is set to $1220 \times 370$ and $1280 \times 384$ for VoxFormer and StereoScene, respectively. 
Other training settings are keep the same as the corresponding baselines. The number of selected hard voxels during training ($N$) is set to $4096$. 
The coefficients ($\alpha$, $\beta$) of linear transformation from $\mathcal{A}$ to $\mathcal{H}^{\text{local}}$ are set to $0.2$ and $1.0$, respectively. The distillation weight $\lambda$ is set to $48$. The trade-off coefficient $\delta$ is set to $0.1$.

All the models in experiments are trained on four GeForce RTX 4090 GPUs with 24G memory, and the inference speed is reported by a single GeForce RTX 4090 GPU. 
More implementation details with different baselines are given in our supplementary material.

\begin{figure*}
\centering
\includegraphics[width=0.95\linewidth]{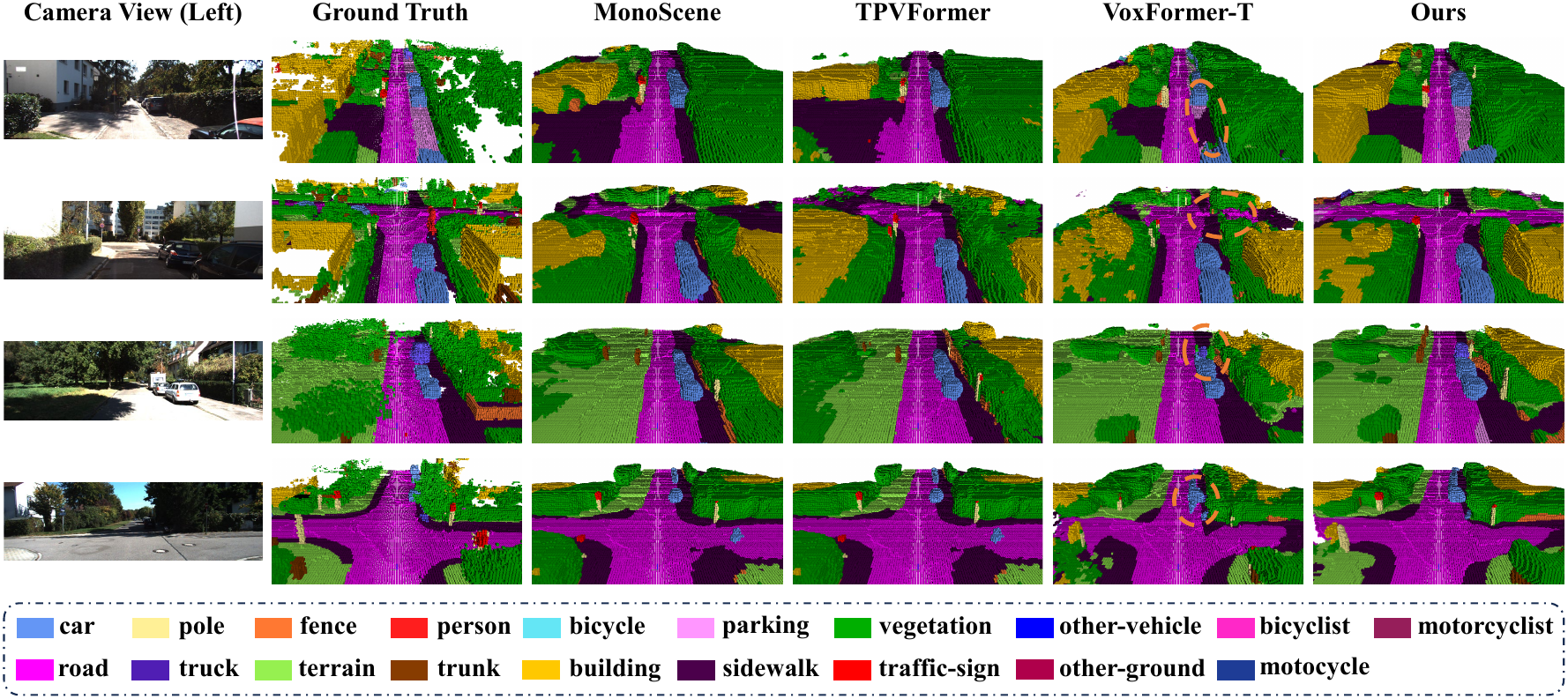}
\vspace{-2mm}
    \caption{Visual results of our method (\textbf{\texttt{HASSC}}-VoxFormer-T) and the state-of-the-art camera-based methods on the validation set of SemanticKITTI. The left shows the perspective view image from left camera, which is the input for model training and inference. The right is the ground truth and the corresponding predicted semantic scene from these methods.}
    \label{fig:vis}
    \vspace{-4mm}
\end{figure*}

\subsection{Performance}
\noindent \textbf{Qualitative Comparisons.} 
We firstly present the quantitative comparison with the our baseline models on the validation set of SemanticKITTI. 
As shown in Tab.~\ref{tab:comp_val}, \textbf{\texttt{HASSC}} effectively improves the accuracy over baseline methods including VoxFormer-S ($+1.13$\%mIoU, $+0.80$\%IoU),  VoxFormer-T ($+1.39$\%mIoU, $+0.43$\%IoU) and StereoScene ($+0.64$\%mIoU, $+0.89$\%IoU) at full range.
Our \textbf{\texttt{HASSC}}-VoxFormer-T obtains a more obvious improvement at the closer range including short ($+2.55$\%mIoU) and middle ($+1.85$\%mIoU) to ensure the safety of autonomous vehicles.
Besides, it is worthy of noting that \textbf{\texttt{HASSC}}-VoxFormer-S with single image input even outperforms VoxFormer-T with $5$ images ($13.48$\% \textit{v.s.} $13.35$\%).

Then, we submit our prediction results to the website of SemanticKITTI for the online evaluation of hidden test set. In Tab.~\ref{tab:hiddentest}, we compare our approach against the state-of-the-art camera-based methods.
\textbf{\texttt{HASSC}} shows consistent improvements with both VoxFormer-S ($+1.14$\%mIoU) and VoxFormer-T ($+0.97$\%mIoU).
Note that \textbf{\texttt{HASSC}}-VoxFormer-S outperforms all the camera-based methods with single image input. Comparisons on detailed semantic categories and further discussions about our method are provided in the supplementary material.

\begin{table}[t]\centering
\footnotesize
\renewcommand\tabcolsep{1.pt}
\renewcommand\arraystretch{1.09}
\begin{tabular}{c|c|c|cc}\toprule
{\textbf{Methods} }  & {\textbf{SSC Input} } & {\textbf{Pub.} } &{\textbf{IoU (\%)}}$\uparrow$ & {\textbf{mIoU (\%)}}$\uparrow$ \\
  \midrule
  \textbf{LMSCNet$^*$}~\cite{roldao2020lmscnet}& $\hat{x}^{\text{occ}}_{\text{3D}}$ & 3DV 2020 & 31.38 & 7.07  \\
\textbf{3DSketch$^*$}~\cite{chen20203d} &  $x^{\text{rgb}}$,$\hat{x}^{\text{TSDF}}$ & CVPR 2020 & 26.85 & 6.23  \\

\textbf{AICNet$^*$}~\cite{li2020anisotropic}& $x^{\text{rgb}}$,$\hat{x}^{\text{depth}}$ & CVPR 2020 & 23.93 & 7.09 \\
\textbf{JS3C-Net$^*$}~\cite{yan2021sparse} & $\hat{x}^{\text{pts}}$ & AAAI 2021 & 34.00 &8.97
\\\midrule
\textbf{MonoScene}~\cite{cao2022monoscene} & $x^{\text{rgb}}$ & CVPR 2022 & 34.16 & 11.08  \\
\textbf{TPVFormer}~\cite{huang2023tri} & $x^{\text{rgb}}$ & CVPR 2023 & 34.25 & 11.26  \\
\textbf{OccFormer}~\cite{zhang2023occformer} & $x^{\text{rgb}}$ & ICCV 2023 & 34.53 & 12.32  \\

\textbf{NDC-Scene}~\cite{yao2023ndc} & $x^{\text{rgb}}$ & ICCV 2023 & 36.19 & 12.58  \\

\textbf{VoxFormer-S}~\cite{li2023voxformer} & $x^{\text{rgb}}$ & CVPR 2023 & 42.95 & 12.20 \\
\textbf{VoxFormer-T}~\cite{li2023voxformer} & $x^{\text{rgb}} \times 5$ & CVPR 2023 & 43.21 & 13.41 \\\midrule

{\textbf{\textbf{\makecell[c]{\texttt{HASSC}-VoxFormer-S}}}} & $x^{\text{rgb}}$ & - & \textbf{43.40} & 13.34 \\

{\textbf{\textbf{\makecell[c]{\texttt{HASSC}-VoxFormer-T}}}} & $x^{\text{rgb}} \times 5$ & - & 42.87 & \textbf{14.38} \\
\bottomrule
\end{tabular}
\vspace{-2mm}
\caption{Quantitative comparisons with the state-of-the-art camera-based methods on the \textbf{hidden test set} of SemanticKITTI. $*$ denotes that the method is converted to the camera-based model by MonoScene~\cite{cao2022monoscene}.}
\label{tab:hiddentest}
\vspace{-6mm}
\end{table}

\noindent \textbf{Qualitative Comparisons.}
To further investigate the effectiveness of our proposed \textbf{\texttt{HASSC}}, we visualize the predictions of different models on the validation set of SemanticKITTI. 
As shown in Fig.~\ref{fig:vis}, our method (\textbf{\texttt{HASSC}}-VoxFormer-T) performs better at complex classes junctions (\textit{e.g.}, road, sidewalk and truck) compared to other camera-based approaches.
This also corresponds to the improvements in quantitative evaluation (road $+3.66$\%mIoU, sidewalk $+2.56$\%mIoU, truck $+9.80$\%mIoU) of our method, which demonstrates the effectiveness of the proposed hard voxel mining scheme.

\subsection{Ablation Studies}
In this section, we perform exhaustive ablation experiments on the validation set of SemanticKITTI with VoxFormer-T~\cite{li2023voxformer} as the baseline model for fair comparison. 

\begin{figure}
\centering
\includegraphics[width=0.95 \linewidth]{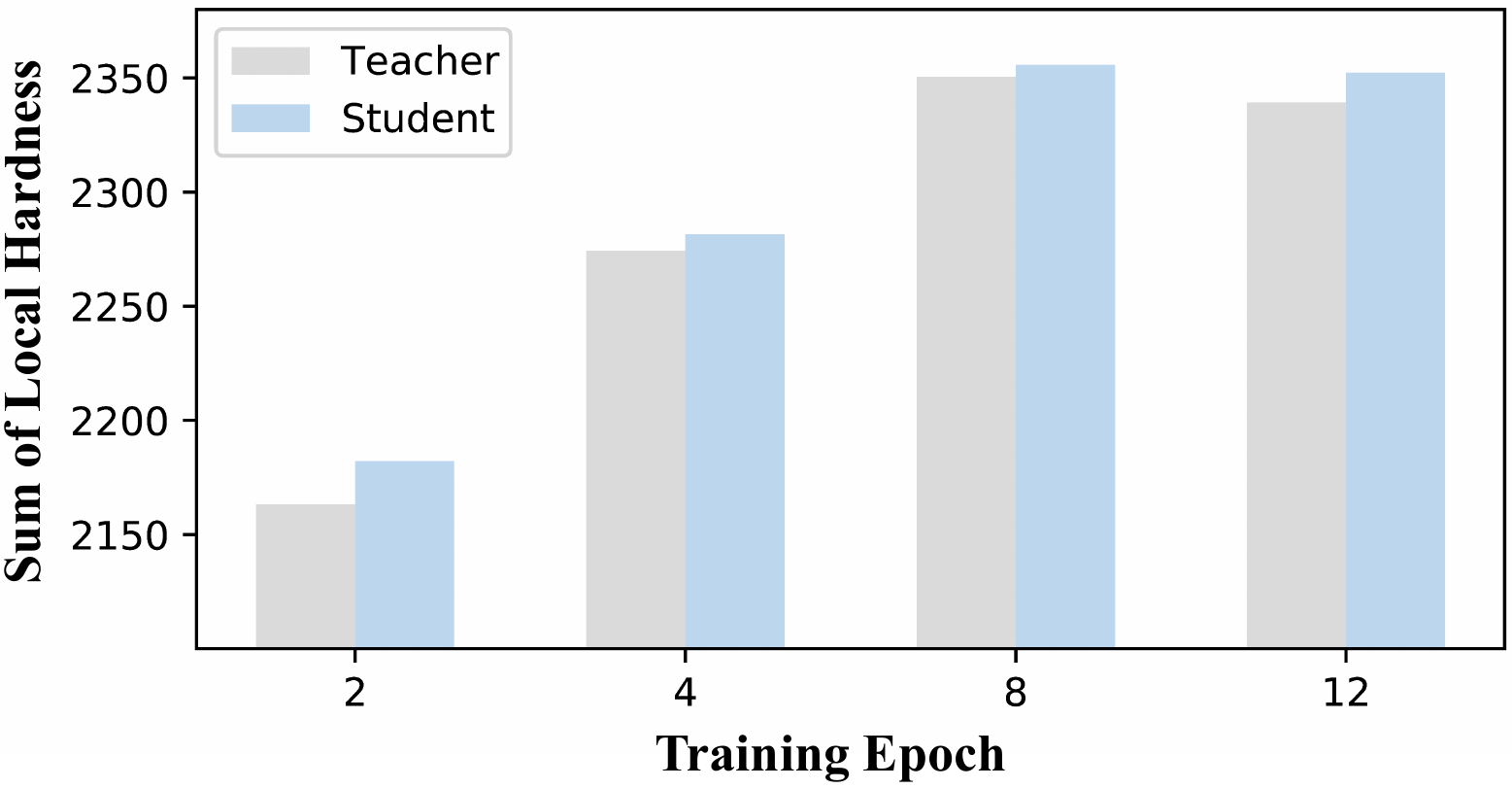}
\vspace{-2mm}
    \caption{Visualization of the sum of the local hardness change during training on both student and teacher branches.}
    \label{fig:training_lh}
    \vspace{-2mm}
\end{figure}

\noindent \textbf{Ablation on \texttt{HASSC} Scheme.}
Firstly, we provide the ablation of the proposed  \textbf{\texttt{HASSC}} scheme.
As illustrated in Tab.~\ref{tab:ablation_all}, the first row is the result reproduced with the original implementation of VoxFormer-T. 
It can be observed that using global hardness $\mathcal{H}^{\text{global}}$ and local hardness $\mathcal{H}^{\text{local}}$ individually obtains the limited performance improvements.
Only the combination of  $\mathcal{H}^{\text{global}}$ and $\mathcal{H}^{\text{local}}$ can effectively improve model performance. 
With the teacher-guided hard voxel mining (T-HVM), \textbf{\texttt{HASSC}} achieves stable improvements in both semantics and geometry.
The self-distillation (T-Distill) from teacher-branch can provide consistent supervision with reliable soft labels and further improve the model accuracy when coupled with HVM head.
Furthermore, we visualize the sum of the local hardness of $N$ selected voxels during training. 
As shown in Fig.~\ref{fig:training_lh}, the sum of local hardness continues to increase on both the student and teacher branches, which means the process of selecting voxels based on the learned global hardness $\mathcal{H}^{\text{global}}$ is consistent with local one $\mathcal{H}^{\text{local}}$.

\begin{table}
    \footnotesize
    \centering
    \renewcommand\tabcolsep{4pt}
    
    \begin{tabular}[b]{cccc|cc}
        \toprule
        \textbf{Global} & \textbf{Local} & \textbf{T-HVM} & \textbf{T-Distill}  & \textbf{IoU (\%)}$\uparrow$ & \textbf{mIoU (\%)}$\uparrow$\\
        \midrule
        & & & &   44.16 & 13.33  \\
        \cmark & & & &  43.89  & 13.30 \\
          & \cmark & & &  44.00  & 13.40 \\
        \cmark & \cmark & & &  43.98  & 13.91 \\
        \cmark & \cmark & \cmark &   & 44.12 & 14.03 \\
         &  &  & \cmark  & 44.38 & 13.65 \\
        \cmark & \cmark & \cmark & \cmark &  \textbf{44.58}  & \textbf{14.74} \\  
        \bottomrule
    \end{tabular}
    \vspace{-2mm}
    \caption{Ablation study on our proposed  \textbf{\texttt{HASSC}} scheme. }
    \label{tab:ablation_all}
    \vspace{-2mm}
\end{table}

\noindent \textbf{Comparison of Training and Inference Efficiency.}
The model complexity analysis regarding training and inference is provided in Tab.~\ref{tab:ablation_head}. 
Compared with the vanilla VoxFormer-T, it can be seen that our proposed \textbf{\texttt{HASSC}} only introduces minimal overheads (+0.90\% parameters) during training but promotes 10.58\% relative performance without incurring extra cost for inference (724.05ms \textit{v.s.} 720.84ms).

\begin{table}
    \footnotesize
    \centering
    \renewcommand\tabcolsep{3.5pt}
    \begin{tabular}[b]{c|cc}
        \toprule
         \textbf{Methods} & \textbf{VoxFormer-T} & \textbf{\texttt{HASSC}-VoxFormer-T}  \\
        \midrule
        \textbf{Params (M)} & 57.91 & 58.43 \\
        \textbf{Inference Speed (ms)} & 724.05 & 720.84 \\
        \textbf{IoU (\%)}$\uparrow$ & 44.16 & \textbf{44.58} \\
        \textbf{mIoU (\%)}$\uparrow$ & 13.33 & \textbf{14.74}  \\
        \bottomrule
    \end{tabular}
    \vspace{-2mm}
    \caption{Comparison with baseline model on the training and inference efficiency. }
    \label{tab:ablation_head}
    \vspace{-5mm}
\end{table}

\noindent \textbf{Ablation on Hard Voxel Selection.}
As there are totally $262,144$ ($128 \times 128\times 16$) voxels in coarse prediction $\mathcal{P}_{\text{coarse}}$, we need to find an appropriate number ($N$) of hard voxels for refinement.
Therefore, the ablation experiments on the number ($N$) of hard voxel selection are shown in Tab.~\ref{tab:voxel_selection}.
These experiments are conducted without distillation loss $\mathcal{L}_{\text{distill}}$ from teacher-branch.
A small $N$ yields an minimal improvement
while a large one may include relatively easy voxels resulting in erroneous optimization.
When $N$ is set to $4096$, the HVM head achieves the best performance.

\noindent \textbf{Ablation on Self-Distillation.}
We present ablation on the weight ($\lambda$) of self-distillation loss $\mathcal{L}_{\text{distill}}$. 
As illustrated in Tab.~\ref{tab:self_distill}, inappropriate distillation loss weight may lead to the inferior model performance when $\lambda=12$. We set $\lambda$ to $48$ in order to better integrate it into the hard voxel mining head.

\begin{table}
    \footnotesize
    \centering
    \renewcommand\tabcolsep{5pt}
    \begin{tabular}[b]{c|ccccc}
        \toprule
          \textbf{Voxel Numbers ($N$)} & 0 & 1024 & 2048& 4096 & 8192  \\
        \midrule
         \textbf{IoU (\%)}$\uparrow$  & \textbf{44.16} & 44.01 & 43.92 & {44.12} & 44.09  \\
         \textbf{mIoU (\%)}$\uparrow$ & 13.33& 13.52 & 13.64 & \textbf{14.03} & 13.74  \\
        \bottomrule
    \end{tabular}
    \vspace{-2mm}
    \caption{Ablation study on the number of  hard voxel selection. }
    \vspace{-2mm}
    \label{tab:voxel_selection}
\end{table}

\begin{table}
    \footnotesize
    \centering
    \renewcommand\tabcolsep{5.5pt}
    \begin{tabular}[b]{c|ccccc}
        \toprule
          \textbf{Distill Weight ($\lambda$)} & 0 &  12 & 24 & 48 & 96  \\
        \midrule
         \textbf{IoU (\%)}$\uparrow$ & 44.12  & 44.06 & 44.25 &  \textbf{44.58}  & 44.51  \\
         \textbf{mIoU (\%)}$\uparrow$ & 14.03 & 13.80 & 14.23 & \textbf{14.74} & 14.39 \\
        \bottomrule
    \end{tabular}
    \vspace{-2mm}
    \caption{Ablation study on the weight of self-distillation from teacher model. }
    \vspace{-2mm}
    \label{tab:self_distill}
\end{table}

\noindent \textbf{Comparison with Other Schemes.}
Finally, we provide comprehensive comparison with existing hard sample mining schemes. 
We re-implement {PALNet}~\cite{li2019depth}, {PointRend}~\cite{kirillov2020pointrend} and~\cite{xiao2023not} with VoxFormer-T~\cite{li2023voxformer} to facilitate a fair comparison.
{PALNet}~\cite{li2019depth} is originally for indoor semantic scene completion and only considers local geometry.
{PointRend}~\cite{kirillov2020pointrend} and~\cite{xiao2023not} are designed for 2D image segmentation and just use the globally updated information from the network optimization process. 
As shown in Tab.~\ref{tab:ablation_compare}, they obtain marginal performance enhancements in 3D space of large-scale scenes. Our proposed \textbf{\texttt{HASSC}} outperforms all the reference methods.

\begin{table}
    \footnotesize
    \centering
    \renewcommand\tabcolsep{4,5pt}
    \begin{tabular}[b]{cc|cc}
        \toprule
        \textbf{Methods} & \textbf{Hardness} & \textbf{IoU (\%)}$\uparrow$ & \textbf{mIoU (\%)}$\uparrow$\\
        \midrule
        \textbf{PALNet}~\cite{li2019depth} & Local  & 44.28 & 13.28\\ \textbf{PointRend}~\cite{kirillov2020pointrend} &  Global & 44.29  & 13.57  \\
        \textbf{Xiao~\textit{et al.}}~\cite{xiao2023not} &  Global & 44.10 & 13.33 \\
        \textbf{Ours} & Global \& Local & \textbf{44.58} & \textbf{14.74} \\
        \bottomrule
    \end{tabular}
    \vspace{-2mm}
    \caption{Comparison with other hard sample mining schemes. We re-implement them with VoxFormer-T for fair comparison.}
    \label{tab:ablation_compare}
    \vspace{-5mm}
\end{table}

\section{Conclusion}
\label{sec:conclusion}
In this paper, we adhere to the principle of \textit{not all voxels are equal} and propose hardness-aware semantic scene completion (\textbf{\texttt{HASSC}}). 
The hard voxel mining head consists of hard voxel selection and  
voxel-wise refinement module, which combines global and local hardness to optimize the network on difficult regions.
Additionally, self-distillation training strategy is introduced to improve the stability and consistency of completion. 
We have conducted extensive experiments to demonstrate that \textbf{\texttt{HASSC}} can effectively promote the existing semantic scene completion models without incurring the overheads during inference.

\section*{Acknowledgments}
This work is supported by National Natural Science Foundation of China under Grants (62376244). It is also supported by Information Technology Center and State Key Lab of CAD\&CG, Zhejiang University.

{
    \small
    \bibliographystyle{ieeenat_fullname}
    \bibliography{main}
}

\makeatletter
\newcommand{\car@semkitfreq}{3.92}
\newcommand{\bicycle@semkitfreq}{0.03}
\newcommand{\motorcycle@semkitfreq}{0.03}
\newcommand{\truck@semkitfreq}{0.16}
\newcommand{\othervehicle@semkitfreq}{0.20}
\newcommand{\person@semkitfreq}{0.07}
\newcommand{\bicyclist@semkitfreq}{0.07}
\newcommand{\motorcyclist@semkitfreq}{0.05}
\newcommand{\road@semkitfreq}{15.30}  %
\newcommand{\parking@semkitfreq}{1.12}
\newcommand{\sidewalk@semkitfreq}{11.13}  %
\newcommand{\otherground@semkitfreq}{0.56}
\newcommand{\building@semkitfreq}{14.1}  %
\newcommand{\fence@semkitfreq}{3.90}
\newcommand{\vegetation@semkitfreq}{39.3}  %
\newcommand{\trunk@semkitfreq}{0.51}
\newcommand{\terrain@semkitfreq}{9.17} %
\newcommand{\pole@semkitfreq}{0.29}
\newcommand{\trafficsign@semkitfreq}{0.08}
\newcommand{\semkitfreq}[1]{{\csname #1@semkitfreq\endcsname}}

\clearpage
\appendix
\section*{Appendix}

\setcounter{figure}{0}
\setcounter{table}{0}
\renewcommand{\thefigure}{A\arabic{figure}}
\renewcommand{\thetable}{A\arabic{table}}

In this supplemental document, we further provide the following descriptions and experiments:

\begin{itemize}
    \item Section \ref{sec:supp_a}: More implementation details;
    \item Section \ref{sec:supp_b}: Additional experiments;
    \item Section \ref{sec:supp_c}: Further discussions. 

\end{itemize}

\section{More Implementation Details}
\label{sec:supp_a}
\noindent \textbf{VoxFormer-based Implementation.}
Following the original settings in VoxFormer~\cite{li2023voxformer}, we adopt monocular images from the left camera and estimated coarse geometry from stereo images as inputs. 
The commonly used ResNet50 backbone~\cite{he2016deep} is employed to extract image features. The 2D-to-3D transformation module consists of three deformable cross-attention layers, while the 3D backbone comprises two deformable self-attention layers. 
The resolution ($X' \times Y' \times Z'$) of 3D fine-grained feature ($\mathcal{F}_{\text{fine}}^{\text{3D}}$) is set to $128 \times 128 \times 16$ and the feature dimension $D$ is set to $128$. 
We train both \textbf{\texttt{HASSC}}-VoxFormer-S and \textbf{\texttt{HASSC}}-VoxFormer-T for $24$ epochs with a learning rate of $2\times10^{-4}$ and a batch size of one sample per GPU.

\noindent \textbf{StereoScene-based Implementation.}
Our \textbf{\texttt{HASSC}}-StereoScene employs stereo images from both the left and right cameras as its inputs. 
We adopt EfficientNet-B7~\cite{tan2019efficientnet} as the image encoder for fair comparison with StereoScene~\cite{li2023stereoscene}.
The 2D-to-3D transformation consists of BEV constructor and stereo constructor with an auxiliary sparse depth supervision from LiDAR point clouds like BEVDepth~\cite{li2023bevdepth}. 
Following MonoScene~\cite{cao2022monoscene}, an encoder-decoder 3D U-Net is used as 3D backbone to process voxel volume features.
The resolution of 3D feature map is set to $128 \times 128 \times 16$. 
Both StereoScene$^{\dag}$ and \textbf{\texttt{HASSC}}-StereoScene in our main paper are trained with 30 epochs, where the learning rate is set to $1\times10^{-4}$ and the batch size is set to $1$ per GPU.

\section{Additional Experiments}
\label{sec:supp_b}
\subsection{Quantitative Comparison}
\noindent \textbf{Comparison with LiDAR-based Methods.} 
We provide a comparison with existing LiDAR-based methods at different ranges. 
As shown in Tab.~\ref{tab:comp2lidar}, \textbf{\texttt{HASSC}}-VoxFormer-T exhibits superior performance, even surpassing several LiDAR-based methods including LMSCNet~\cite{roldao2020lmscnet} and SSCNet~\cite{song2017semantic} ($24.10$\%mIoU \textit{v.s.} $22.37$\%mIoU / $20.02$\%mIoU) at short range. This outcome underscores the effectiveness of our approach in improving model accuracy.
\begin{table}[ht]\centering
\footnotesize
\renewcommand\tabcolsep{1.pt}
\renewcommand\arraystretch{1.09}
\begin{tabular}{c|c|ccc|ccc}\toprule
\multirow{2}{*}{\textbf{Methods} }  & \multirow{2}{*}{\textbf{Modality} } &\multicolumn{3}{c|}{\textbf{IoU (\%)}$\uparrow$} &\multicolumn{3}{c}{\textbf{mIoU (\%)}$\uparrow$} \\
  &  & \textbf{S} & \textbf{M} & \textbf{L} & \textbf{S} & \textbf{M} & \textbf{L} \\\midrule
\textbf{JS3CNet$^*$}~\cite{yan2021sparse} & LiDAR &63.47 & {63.40} & {53.09} & \textbf{30.55} & \textbf{28.12} & \textbf{22.67} \\
\textbf{LMSCNet$^*$}~\cite{roldao2020lmscnet} & LiDAR & \textbf{74.88} & \textbf{69.45} & \textbf{55.22} & {22.37} & {21.50} & {17.19} \\
\textbf{SSCNet$^*$}~\cite{song2017semantic} & LiDAR &{64.37} & {61.02} & {50.22} &20.02 & {19.68} & {16.35} \\
\midrule
\textbf{\texttt{HASSC}-VoxFormer-T}& Camera & \textbf{66.05} & \textbf{58.01} & \textbf{44.58} & \textbf{24.10} & \textbf{20.27} & \textbf{14.74} \\
\textbf{VoxFormer-T}~\cite{li2023voxformer}& Camera & {65.38} &57.69 &44.15 & {21.55} &18.42 &13.35 \\
\textbf{TPVFormer}~\cite{huang2023tri} & Camera & {54.75} & 46.03 & 35.62 & {17.15} & 15.27 & 11.30 \\
\textbf{MonoScene$^*$}~\cite{cao2022monoscene} & Camera & 38.42 & 38.55 & 36.80 & 12.25 & 12.22 & 11.30 \\

\bottomrule
\end{tabular}
\vspace{-2mm}
\caption{Quantitative comparison with the existing LiDAR-based semantic scene completion methods. 
$*$ denotes that the results are reported in VoxFormer~\cite{li2023voxformer}.}
\label{tab:comp2lidar}
\vspace{-3mm}
\end{table}

\noindent \textbf{More Ablation Studies.}
Firstly, we conduct ablation experiments on the coefficients ($\alpha$, $\beta$) of linear transformation from $\mathcal{A}$ to $\mathcal{H}^{\text{local}}$. 
The results, as shown in Table~\ref{tab:ablation_ab}, indicate that the value of $\alpha$ has a greater impact on the completion accuracy than $\beta$.
We set $\alpha=0.2$ and $\beta=1.0$ empirically.

\begin{table}[t]
    \footnotesize
    \centering
    \renewcommand\tabcolsep{12pt}
    \begin{tabular}[b]{cc|cc}
        \toprule
        \textbf{$\alpha$} & \textbf{$\beta$} & \textbf{IoU (\%)}$\uparrow$ & \textbf{mIoU (\%)}$\uparrow$\\
        \midrule
        0.1 & 1.0  &  \textbf{44.66} & 14.40 \\ 
        0.2 & 1.0 & {44.58} & \textbf{14.74}  \\
        0.4 & 1.0 &  44.46 & 14.34  \\
        0.2 & 0.8 & 44.64  & 14.71   \\
        0.2 & 1.2 &  44.48 &  14.47  \\
        \bottomrule
    \end{tabular}
    \vspace{-2mm}
    \caption{Ablation study on the coefficients ($\alpha$, $\beta$) of linear transformation from $\mathcal{A}$ to $\mathcal{H}^{\text{local}}$.}
    \label{tab:ablation_ab}
\vspace{-3mm}
\end{table}

\begin{table}[t]
    \footnotesize
    \centering
    \renewcommand\tabcolsep{12pt}
    \begin{tabular}[b]{cc|cc}
        \toprule
        \textbf{$t$} & \textbf{$\omega$} & \textbf{IoU (\%)}$\uparrow$ & \textbf{mIoU (\%)}$\uparrow$\\
        \midrule
        
        1 & 0.75  & 44.52  & 14.25 \\ 
        3 & 0.75 & {44.58} & \textbf{14.74}  \\
        5 & 0.75 & \textbf{44.61}  & 14.57  \\
        10 & 0.75 & 44.33  & 14.04  \\
        3 & 0.5 &  44.46 & 14.61   \\
        3 & 1.0 &  44.51 &  14.69  \\
        1 & 1.0  &  44.53  & 14.50  \\ 
        \bottomrule
    \end{tabular}
    \vspace{-2mm}
    \caption{Ablation study on the hyper-parameters ($t$, $\omega$) for hard voxel selection strategy.}
    \label{tab:ablation_tw}
    \vspace{-6mm}
\end{table}

\begin{table*}[htb]
\renewcommand\tabcolsep{1.6pt}
\footnotesize
\newcommand{\classfreq}[1]{{~\scriptsize(\semkitfreq{#1}\%)}} 
    \centering
    \begin{tabular}{r|c|ccccccccccccccccccc|c}
   \toprule 
    & \textbf{SC} & \multicolumn{20}{c}{\textbf{SSC}} \\
\multicolumn{1}{c|}{\textbf{Methods}} &\rotatebox{90}{\textbf{IoU (\%)}}
& \rotatebox{90}{\crule[carcolor]{0.18cm}{0.18cm} \textbf{car}\classfreq{car}}
& \rotatebox{90}{\crule[bicyclecolor]{0.18cm}{0.18cm} \textbf{bicycle}\classfreq{bicycle}}
& \rotatebox{90}{\crule[motorcyclecolor]{0.18cm}{0.18cm} \textbf{motorcycle}\classfreq{motorcycle}} 
& \rotatebox{90}{\crule[truckcolor]{0.18cm}{0.18cm} \textbf{truck}\classfreq{truck}} 
& \rotatebox{90}{\crule[othervehiclecolor]{0.18cm}{0.18cm} \textbf{other-veh.}\classfreq{othervehicle}}
& \rotatebox{90}{\crule[personcolor]{0.18cm}{0.18cm} \textbf{person}\classfreq{person}}
& \rotatebox{90}{\crule[bicyclistcolor]{0.18cm}{0.18cm} \textbf{bicyclist}\classfreq{bicyclist}}
& \rotatebox{90}{\crule[motorcyclistcolor]{0.18cm}{0.18cm} \textbf{motorcyclist}\classfreq{motorcyclist}}
& \rotatebox{90}{\crule[roadcolor]{0.18cm}{0.18cm} \textbf{road}\classfreq{road}}
& \rotatebox{90}{\crule[parkingcolor]{0.18cm}{0.18cm} \textbf{parking}\classfreq{parking}} 
& \rotatebox{90}{\crule[sidewalkcolor]{0.18cm}{0.18cm} \textbf{sidewalk}\classfreq{sidewalk}}
& \rotatebox{90}{\crule[othergroundcolor]{0.18cm}{0.18cm} \textbf{other-grnd}\classfreq{otherground}}
& \rotatebox{90}{\crule[buildingcolor]{0.18cm}{0.18cm} \textbf{building}\classfreq{building}}
& \rotatebox{90}{\crule[fencecolor]{0.18cm}{0.18cm} \textbf{fence}\classfreq{fence}}
& \rotatebox{90}{\crule[vegetationcolor]{0.18cm}{0.18cm} \textbf{vegetation}\classfreq{vegetation}}
& \rotatebox{90}{\crule[trunkcolor]{0.18cm}{0.18cm} \textbf{trunk}\classfreq{trunk}}
&\rotatebox{90}{\crule[terraincolor]{0.18cm}{0.18cm} \textbf{terrain}\classfreq{terrain}}
&\rotatebox{90}{\crule[polecolor]{0.18cm}{0.18cm} \textbf{pole}\classfreq{pole}} 
&\rotatebox{90}{\crule[trafficsigncolor]{0.18cm}{0.18cm} \textbf{traf.-sign}\classfreq{trafficsign}}
&\rotatebox{90}{\textbf{mIoU (\%)}}
\\ \midrule

\textbf{LMSCNet$^*$}~\pub{3DV20}~\cite{roldao2020lmscnet} & 31.38 & 14.30 & 0.00 &  0.00 & 0.30 & 0.00 & 0.00 & 0.00 &  0.00 & 46.70 & 13.50 & 19.50 & 3.10 & 10.30 & 5.40 & 10.80 & 0.00 & 10.40 & 0.00 & 0.00 & 7.07   \\
\textbf{3DSketch$^*$}~\pub{CVPR20}~\cite{chen20203d} & 26.85 & 17.10 & 0.00 & 0.00 & 0.00 & 0.00 & 0.00 & 0.00 & 0.00 & 37.70 & 0.00 & 19.80 &  0.00 & 12.10 & 3.40 & 12.10 & 0.00 & 16.10 & 0.00 & 0.00 & 6.23  \\
\textbf{AICNet$^*$}~\pub{CVPR20}~\cite{li2020anisotropic} & 23.93 & 15.30 & 0.00 & 0.00 & 0.70 & 0.00 & 0.00 & 0.00 & 0.00 & 39.30 & 19.80 & 18.30 & 1.60 & 9.60 & 5.00 & 9.60 & 1.90 & 13.50 & 0.10 & 0.00 &  7.09 \\
\textbf{JS3C-Net$^*$}~\pub{AAAI21}~\cite{yan2021sparse} & 34.00 & 20.10 & 0.00 & 0.00 & 0.80 & 4.10 & 0.00 & 0.20 & 0.20 & 47.30 & 19.90 & 21.70 & 2.80 & 12.70 & 8.70 & 14.20 & 3.10 & 12.40 & 1.90 & 0.30 &  8.97 \\
\textbf{MonoScene}~\pub{CVPR22}~\cite{cao2022monoscene} & 34.16 & 18.80 & 0.50 & 0.70 & {3.30} & {{4.40}} & {1.00} & 1.40 & \underline{0.40}& {54.70}& {24.80}& {27.10}& 5.70& 14.40& {11.10}& 14.90&2.40& 19.50& 3.30& 2.10& 11.08 \\
\textbf{TPVFormer}~\pub{CVPR23}~\cite{huang2023tri} & {34.25} & 19.20 & 1.00 & 0.50 & {3.70} & 2.30 & {1.10} & {{2.40}} & 0.30 & 55.10 & \underline{27.40} & 27.20 &{6.50} & 14.80 & 11.00 & 13.90 & 2.60 & 20.40& 2.90 & 1.50 & 11.26\\
\textbf{OccFormer}~\pub{ICCV23}~\cite{zhang2023occformer} & {34.53} & 21.60 & 1.50 & \underline{1.70} & 1.20 & 3.20 & \underline{2.20} & {1.10} & 0.20 & \underline{55.90} & \textbf{31.50} & \textbf{30.30} &{6.50} & 15.70 & 11.90 &16.80 & 3.90& 21.30& 3.80 & 3.70 & 12.32\\
\textbf{NDC-Scene}~\pub{ICCV23}~\cite{yao2023ndc} & {36.19} & 19.13 & \textbf{1.93} & \textbf{2.07} & \textbf{4.77} & \textbf{6.69} & \textbf{{3.44}} & {2.77} & \textbf{1.64} & \textbf{58.12} & {25.31} & {28.05} &{6.53} & 14.90 & 12.85 &17.94 & 3.49 & \underline{25.01} & 4.43 & 2.96 & 12.58\\
\midrule
\textbf{VoxFormer-S}~\pub{CVPR23}~\cite{li2023voxformer} &{42.95} & 20.80 & 1.00 & 0.70 & 3.50 & 3.70 & {1.40} & {2.60} & 0.20 & 53.90 & 21.10 & 25.30 &{5.60} & 19.80 & 11.10 &22.40 & 7.50& 21.30& 5.10 & 4.90 & 12.20\\

\textbf{\texttt{HASSC}-VoxFormer-S} &  \textbf{{43.40}}  &  \underline{22.80} &  {1.60} &  {1.00} &  \underline{4.70} &  {3.90} &  {1.60} &  \textbf{{4.00}} &  {0.30} &  {54.60} &  {23.80} &  {27.70} &  {6.20} &  {21.10} &  \underline{13.10} &  {23.80} &  \underline{8.50} &  {23.30} &  {5.80} &  {5.50} &  {13.34} \\

\midrule
\textbf{VoxFormer-T}~\pub{CVPR23}~\cite{li2023voxformer} & \underline{43.21} & {21.70} & \underline{1.90} & {{1.60}} & {3.60} & 4.10& {1.60} & {1.10}& 0.00& 54.10 & {25.10} & 26.90& \underline{7.30} & \textbf{23.50} & \underline{13.10}& \underline{24.40}& {{8.10}}& {{24.20}}& \underline{6.60}& \underline{5.70}& \underline{13.41}\\

\textbf{\texttt{HASSC}-VoxFormer-T} & 42.87  & \textbf{{23.00}} & \underline{1.90} & 1.50 & 2.90 &  \underline{4.90} & 1.40 & \underline{3.00} & 0.00 & 55.30 & 25.90 & \underline{29.60} & \textbf{11.30} & \underline{23.10} & \textbf{14.30} & \textbf{24.80} & \textbf{9.80} & \textbf{26.50} & \textbf{7.00} & \textbf{7.10} & \textbf{14.38} \\

\bottomrule
\end{tabular}
\vspace{-2mm}
    \caption{Performance comparisons of detailed semantic categories with the state-of-the-art camera-based methods on the \textbf{\textit{hidden test set}} of SemanticKITTI~\cite{behley2019semantickitti}. $*$ denotes that the method is converted to the camera-based model by MonoScene~\cite{cao2022monoscene}. 
    The top two with the highest accuracy are highlighted in \textbf{bold} and \underline{underlined}, respectively.}
    \label{tab:supp_hiddentest}
\vspace{-2mm}
\end{table*}

\begin{table*}[htb]
\renewcommand\tabcolsep{1.6pt}
\footnotesize
\newcommand{\classfreq}[1]{{~\scriptsize(\semkitfreq{#1}\%)}} 
    \centering
    \begin{tabular}{r|c|ccccccccccccccccccc|c}
   \toprule 
    & \textbf{SC} & \multicolumn{20}{c}{\textbf{SSC}} \\
\multicolumn{1}{c|}{\textbf{Methods}} &\rotatebox{90}{\textbf{IoU (\%)}}
& \rotatebox{90}{\crule[carcolor]{0.18cm}{0.18cm} \textbf{car}}
& \rotatebox{90}{\crule[bicyclecolor]{0.18cm}{0.18cm} \textbf{bicycle}}
& \rotatebox{90}{\crule[motorcyclecolor]{0.18cm}{0.18cm} \textbf{motorcycle}} 
& \rotatebox{90}{\crule[truckcolor]{0.18cm}{0.18cm} \textbf{truck}} 
& \rotatebox{90}{\crule[othervehiclecolor]{0.18cm}{0.18cm} \textbf{other-veh.}}
& \rotatebox{90}{\crule[personcolor]{0.18cm}{0.18cm} \textbf{person}}
& \rotatebox{90}{\crule[bicyclistcolor]{0.18cm}{0.18cm} \textbf{bicyclist}}
& \rotatebox{90}{\crule[motorcyclistcolor]{0.18cm}{0.18cm} \textbf{motorcyclist}}
& \rotatebox{90}{\crule[roadcolor]{0.18cm}{0.18cm} \textbf{road}}
& \rotatebox{90}{\crule[parkingcolor]{0.18cm}{0.18cm} \textbf{parking}} 
& \rotatebox{90}{\crule[sidewalkcolor]{0.18cm}{0.18cm} \textbf{sidewalk}}
& \rotatebox{90}{\crule[othergroundcolor]{0.18cm}{0.18cm} \textbf{other-grnd}}
& \rotatebox{90}{\crule[buildingcolor]{0.18cm}{0.18cm} \textbf{building}}
& \rotatebox{90}{\crule[fencecolor]{0.18cm}{0.18cm} \textbf{fence}}
& \rotatebox{90}{\crule[vegetationcolor]{0.18cm}{0.18cm} \textbf{vegetation}}
& \rotatebox{90}{\crule[trunkcolor]{0.18cm}{0.18cm} \textbf{trunk}}
&\rotatebox{90}{\crule[terraincolor]{0.18cm}{0.18cm} \textbf{terrain}}
&\rotatebox{90}{\crule[polecolor]{0.18cm}{0.18cm} \textbf{pole}} 
&\rotatebox{90}{\crule[trafficsigncolor]{0.18cm}{0.18cm} \textbf{traf.-sign}}
&\rotatebox{90}{\textbf{mIoU (\%)}}
\\ \midrule
\textbf{VoxFormer-T} & 44.15 & 26.54 & \textbf{1.28} & 0.56 & 7.26 & 7.81 & 1.93 & 1.97 & 0.00 & 53.57 & 19.69 & 26.52 & 0.42 & 19.54 & 7.31 & 26.10 & 6.10 & 33.06 & 9.15 & 4.94 &  13.35 \\

\textbf{\texttt{HASSC}-VoxFormer-T$^{*}$} & 44.12 & 26.63 & 0.26 & 0.57 & 6.63 & 8.45 & \textbf{2.67} & 2.47 & 0.00 & 56.57 & \textbf{20.81} & 29.04 & 0.77 & 19.89 & 7.87 & \textbf{27.08} & 7.51 & \textbf{34.49} & \textbf{9.42} & \textbf{5.55} & 14.03\\

\textbf{\texttt{HASSC}-VoxFormer-T} & \textbf{44.58}  & \textbf{27.33} & {1.07} & \textbf{1.14} & \textbf{17.06} &  \textbf{8.83} & 2.25 & \textbf{4.09} & 0.00 & \textbf{57.23} & 19.89 & \textbf{29.08} & \textbf{1.26} & \textbf{20.19} & \textbf{7.95} & 27.01 & \textbf{7.71} & 33.95 & 9.20 & 4.81 & \textbf{14.74} \\

\bottomrule
\end{tabular}
\vspace{-2mm}
    \caption{Performance comparisons of detailed semantic categories for ablation on the \textbf{\textit{validation set}} of SemanticKITTI.  $*$ denotes no self-distillation during training.}
    \label{tab:supp_ablclass}
\vspace{-2mm}
\end{table*}

\begin{table*}[!h]
\renewcommand\tabcolsep{4pt}
\footnotesize
\newcommand{\classfreq}[1]{{~\scriptsize(\semkitfreq{#1}\%)}} 
    \centering
    \begin{tabular}{r|c|ccccccccccc|c}
   \toprule 
    & \textbf{SC} & \multicolumn{12}{c}{\textbf{SSC}} \\
\multicolumn{1}{c|}{\textbf{Methods}}&\rotatebox{90}{\textbf{IoU (\%)}}
& \rotatebox{90}{\textbf{person}}
& \rotatebox{90}{\textbf{rider}}
& \rotatebox{90}{\textbf{car}} 
& \rotatebox{90}{\textbf{trunk}} 
& \rotatebox{90}{\textbf{plants}}
& \rotatebox{90}{\textbf{traf.-sign}}
& \rotatebox{90}{\textbf{pole}}
& \rotatebox{90}{\textbf{building}}
& \rotatebox{90}{\textbf{fence}}
& \rotatebox{90}{\textbf{bike}} 
& \rotatebox{90}{\textbf{ground}}

&\rotatebox{90}{\textbf{mIoU (\%)}}
\\ \midrule

\textbf{SSCNet}$^{\dag}$~\pub{CVPR17}~\cite{song2017semantic} & 40.82 & {10.13} & 0.10  & {1.68}  & 2.72 & 31.80 & {1.36} & {6.24} &  32.95 & 4.41  & 15.23 & 25.84 &  12.04 \\

\textbf{\texttt{HASSC}-SSCNet} & \tbblue{42.66} & 8.86 & \tbblue{0.17} & 1.01  & \tbblue{3.82} & \tbblue{34.22} & 0.63 & 6.12 & \tbblue{35.03}  & \tbblue{5.57} & \tbblue{17.04} & 25.84  &  \tbblue{12.57}
\\\midrule

\textbf{LMSCNet}$^{\dag}$~\pub{3DV20}~\cite{roldao2020lmscnet} & 51.97 & 5.60 & 0.00 &  {2.03} & 2.66 & 38.00 & 2.21 & {5.91} & 40.05  & 13.52 & 28.14 &  43.93 &  16.55 \\

\textbf{\texttt{HASSC}-LMSCNet} & \tbblue{52.38} & \tbblue{7.53} & \tbblue{0.36} & 0.66 & \tbblue{3.07} & \tbblue{39.77} & \tbblue{2.83}  & 3.37 & \tbblue{40.47} & \tbblue{15.29}  & \tbblue{32.20} & \tbblue{46.57} &  \tbblue{17.47} \\

\bottomrule
\end{tabular}
\vspace{-2mm}
    \caption{Quantitative comparisons with the LiDAR-based baseline models on the \textbf{\textit{validation set}} of SemanticPOSS~\cite{pan2020semanticposs}. We only use the hard voxel mining (HVM) head \textbf{without} self-training strategy. $\dag$ denotes the results are reproduced from the original implementation. The improved results compared to the corresponding baselines are marked in \tbblue{blue}.}
    \label{tab:supp_poss}
\vspace{-4mm}
\end{table*}

Then, the ablations on the hyper-parameters ($t$, $\omega$) for hard voxel selection are provided in Tab.~\ref{tab:ablation_tw}. 
The settings of $t$ and $\omega$ are crucial for selecting hard voxels to ensure diversity, which can prevent the model from over-fitting in local areas as described in Sec. 3.2 of main paper. As illustrated in Tab.~\ref{tab:ablation_tw}, appropriately expanding the scope of hard voxel selection is beneficial to improving model accuracy.

\noindent \textbf{Comparison on Detailed Semantic Categories on Official Benchmark.}
The detailed semantic categories comparison on the \textit{test set} of SemanticKITTI~\cite{behley2019semantickitti} is illustrated in Tab.~\ref{tab:supp_hiddentest}. \textbf{\texttt{HASSC}}-VoxFormer-T outperforms all  the camera-based methods. 
Notably, \textbf{\texttt{HASSC}}-VoxFormer-S shows obvious improvements in all semantic categories comparing to VoxFormer-S.
We also present class-wise mIoU results on the \textit{validation set} for ablation. As shown in Tab.~\ref{tab:supp_ablclass}, the hard voxel mining yields obvious improvements on \textit{road} (+3.00\% mIoU) and \textit{sidewalk} (+2.52\% mIoU) categories, which possess substantial volumes and regular geometries. The self-distillation combined with HVM head further improves the accuracy on \textit{truck}/\textit{car}/\textit{bicyclist} benefiting from the consistency in geometry.

\begin{figure*}[ht]
\centering
\includegraphics[width=1.0\linewidth]{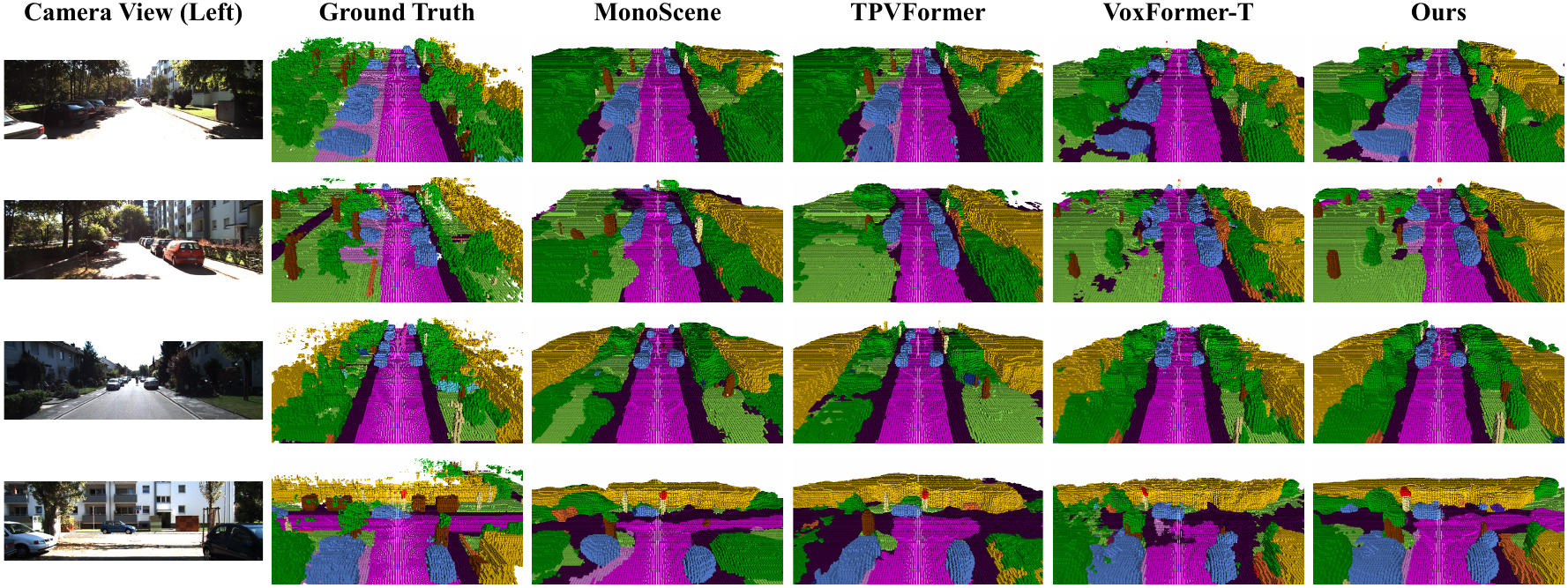}
    \vspace{-5mm}
    \caption{Additional visual results of our method (\textbf{\texttt{HASSC}}-VoxFormer-T) and the state-of-the-art camera-based methods on the \textbf{\textit{validation set}} of SemanticKITTI.}
    \label{fig:supp_vis_val}
     \vspace{-3mm}
\end{figure*}

\begin{figure}[t]
\centering
\includegraphics[width=0.95 \linewidth]{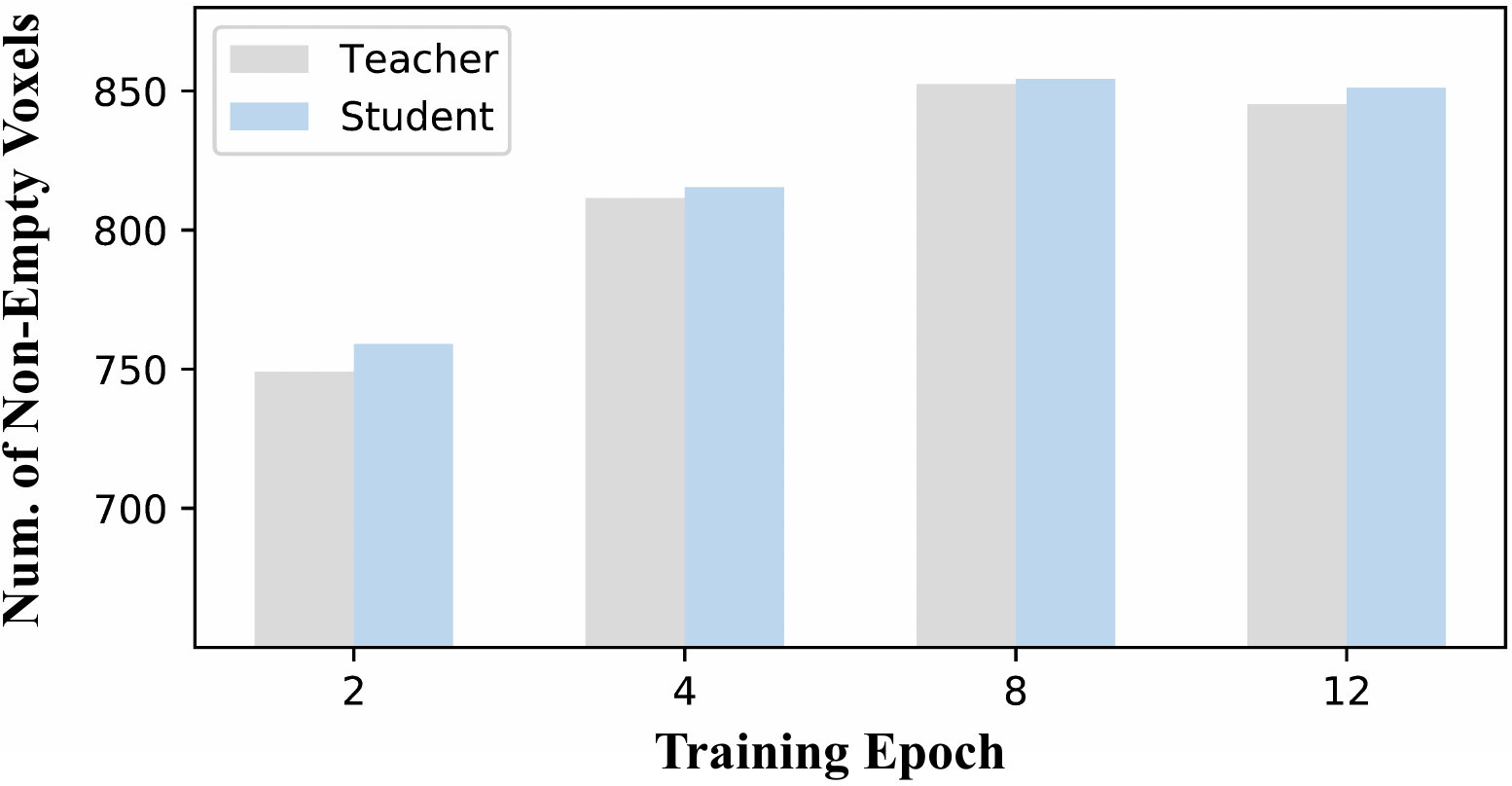}
\vspace{-2mm}
    \caption{The number of the non-empty voxels in selected hard voxels changes during training on both student and teacher branches.}
    \label{fig:training_nonempty}
    \vspace{-4mm}
\end{figure}

\noindent \textbf{More Experiments on Other Dataset.} To examine the scalability of our proposed hard voxel mining (HVM) head, we conduct experiments on another challenging dataset SemanticPOSS~\cite{pan2020semanticposs}. 
Since SemanticPOSS is collected in campus scenarios, there are many moving objects like \textit{person} and \textit{rider} that are hard to identify.
SemanticPOSS contains six sequences and only provides LiDAR point clouds as input. 
We follow the original setting and split the (00-01, 03-05) / 02 as training/validation set.
LiDAR-based methods including SSCNet~\cite{song2017semantic} and LMSCNet~\cite{roldao2020lmscnet} are utilized as baseline models. 
We only replace the vanilla completion head with HVM head and retrain these models. Note that the self-training strategy is not used here.
As shown in Tab.~\ref{tab:supp_poss}, our HVM head achieves stable improvements with various models including SSCNet (+0.53\%mIoU, +1.84\%IoU) and LMSCNet (+0.92\%mIoU, +0.41\%IoU).

\subsection{Qualitative Comparison}
\noindent \textbf{Visualization of Selected Hard Voxels.} 
We visualize the number of the non-empty voxels in selected hard voxels on both student and teacher branches. 
As illustrated in Fig.~\ref{fig:training_nonempty}, the number of the non-empty voxels continues to increase during training, which indicates that our dynamic hard voxel selection strategy favors more difficult non-empty voxels.

\noindent \textbf{Comparison on Validation Set.}
Here, we provide more qualitative comparisons on the \textit{validation set} of SemanticKITTI. As shown in Fig.~\ref{fig:supp_vis_val}, our method (\textbf{\texttt{HASSC}}-VoxFormer-T) obtains stable completion results in complex scenarios compared to other camera-based methods.

\begin{figure*}[ht]
\centering
\includegraphics[width=1.0\linewidth]{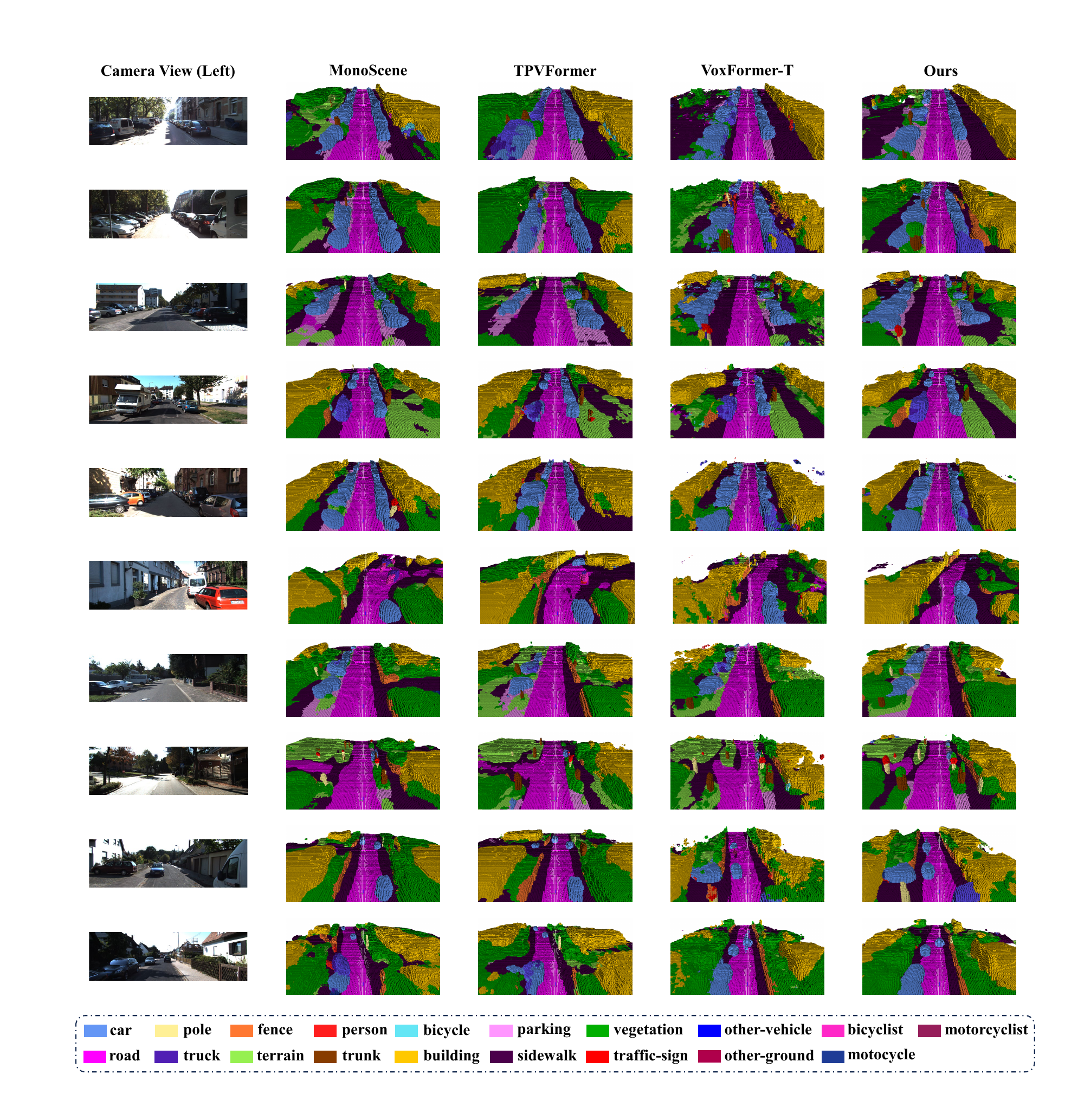}
\vspace{-4mm}
    \caption{Visual results of our method (\textbf{\texttt{HASSC}}-VoxFormer-T) and other camera-based methods on the \textbf{\textit{hidden test set}} of SemanticKITTI.}
    \label{fig:supp_vis_test}
    \vspace{-4mm}
\end{figure*}

\noindent \textbf{Comparison on Test Set.}
Additionally, several qualitative comparisons on the \textit{test set} are presented in Fig.~\ref{fig:supp_vis_test}.
Our method consistently performs well on the more challenging \textit{test set} with 11 sequences.

\section{Further Discussions}
\label{sec:supp_c}

\subsection{Explanatory Experiments}
\begin{figure}[!h]
\centering
\includegraphics[width=0.95 \linewidth]{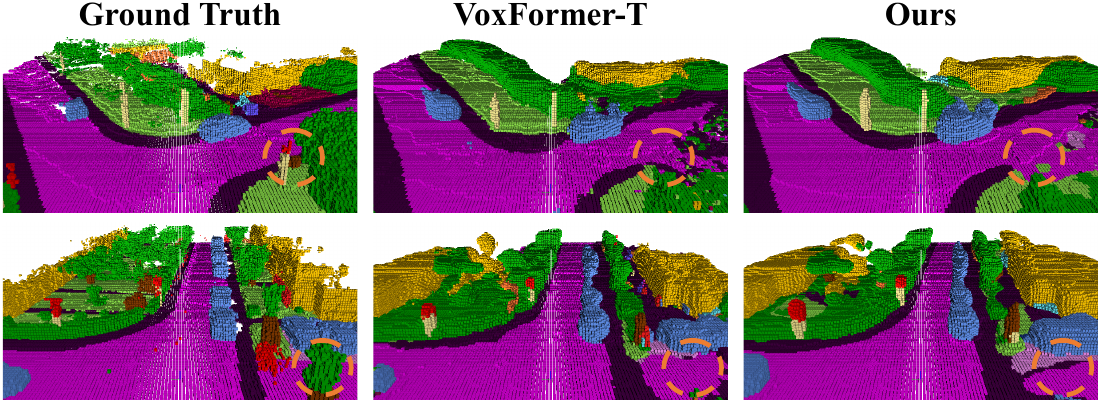}
\vspace{-2mm}
    \caption{Visualization of some typical failure cases.}
    \label{fig:fc}
    \vspace{-2mm}
\end{figure}

\noindent \textbf{Failure Cases.} There are mainly two factors affecting performance of our method: 1) some categories exhibit a long-tail distribution with infrequent occurrence, like \textit{bicyclist} (0.07\%), \textit{traffic sign} (0.08\%) and \textit{trunk} (0.51\%); 2) certain classes, like \textit{vegetation}, often have sever occlusion, which brings in difficulties in geometry estimation. We provide several failure cases caused by the false geometric estimation in the Fig.~\ref{fig:fc}. 

\vspace{-1mm}

\noindent \textbf{About the Formulation of Hardness.} We have explored alternative designs of hardness, such as utilizing the highest confidence (``A'') or loss magnitude (``B'') as indicators for global hardness, and quadratic function encoding for local hardness (``C''). The Tab.~\ref{tab:supp_design} indicates that Eq.1 and Eq.3 of main paper can better integrate global information with local geometric priors.

\begin{table}[h]
    \footnotesize
    \centering
    \renewcommand\tabcolsep{6.4pt}
    \renewcommand\arraystretch{1.09}
    \begin{tabular}[b]{c|cccc}
        \toprule
          \textbf{Method} & \textbf{Design A} & \textbf{Design B} & \textbf{Design C} & \textbf{Ours}  \\
        \midrule
         \textbf{mIoU (\%)}$\uparrow$ & 14.21 & 14.01 & 14.31 & \textbf{14.74} \\
        \bottomrule
    \end{tabular}
    \vspace{-2mm}
     \caption{Comparison of different design options.}
     \vspace{-2mm}
    \label{tab:supp_design}
\end{table}

\noindent \textbf{About the Impact of Empty Voxels.} Empty voxels, typically more numerous, are simpler to classify.
We leverage prior ratio information of empty/non-empty voxels with various hardness levels for hard voxel weighting and model training to enhance voxel-wise SSC.
As shown in the Tab.~\ref{tab:supp_training},
\textbf{\texttt{HASSC}} effectively accelerates network convergence and attains promising accuracy. 

\begin{table}[h]
    \footnotesize
    \centering
    \renewcommand\tabcolsep{1.0pt}
    \renewcommand\arraystretch{1.09}
    \vspace{-1mm}
    \scalebox{1.0}{
    \begin{tabular}[b]{c|ccccc|c}
        \toprule
          \textbf{Epoch} & 2 & 4 & 6 & 8 & 12 & Final (best)  \\
        \midrule
         \textbf{VoxFormer-T} & 10.57 & 11.91 & 11.95& 12.47 & 13.06 &13.33  \\
         \textbf{\texttt{HASSC}-VoxFormer-T} & \textbf{10.95} & \textbf{12.37}  & \textbf{13.66} & \textbf{14.20}  & \textbf{14.74} & \textbf{14.74}     \\
        \bottomrule
    \end{tabular}}
    \vspace{-2mm}
    \caption{Illustration of accuracy variations during training.}
    \vspace{-4mm}
    \label{tab:supp_training}
\end{table}

\subsection{Limitation and Future Work}

Although having achieved promising improvements over existing methods especially at short/middle range with our proposed scheme, there is still a large performance gap between camera-based methods and LiDAR-based methods~\cite{cheng2021s3cnet, xia2023scpnet} in the full range. 
Besides, over-fitting on hard voxels is indeed a potential problem. Our method is also constrained by the inaccurate geometry estimation and the long-tail distribution. 

In the future, we will leverage neural radiance fields (NeRFs)~\cite{mildenhall2021nerf} to extract the geometric and semantic clues contained in the image sequences to further improve the performance of \textit{vision-centric} methods.

\end{document}